# Intra-Retinal Layer Segmentation of 3D Optical Coherence Tomography Using Coarse Grained Diffusion Map

Raheleh Kafieh[1], Hossein Rabbani[*1,2], *Member, IEEE*, Michael D. Abramoff, *Senior Member, IEEE*, Milan Sonka, *Fellow, IEEE*

[1]Department of Physics and Biomedical Engineering, Medical Image and Signal Processing Research Center, Isfahan University of Medical Sciences, Isfahan, IRAN, email: r_kafieh@yahoo.com

*Abstract*—Optical coherence tomography (OCT) is a powerful and noninvasive method for retinal imaging. In this paper, we introduce a fast segmentation method based on a new variant of spectral graph theory named diffusion maps. The research is performed on spectral domain (SD) OCT images depicting macular and optic nerve head appearance. The presented approach does not require edge-based image information and relies on regional image texture. Consequently, the proposed method demonstrates robustness in situations of low image contrast or poor layer-to-layer image gradients. Diffusion mapping is applied to 2D and 3D OCT datasets composed of two steps, one for partitioning the data into important and less important sections, and another one for localization of internal layers. In the first step, the pixels/voxels are grouped in rectangular/cubic sets to form a graph node. The weights of a graph are calculated based on geometric distances between pixels/voxels and differences of their mean intensity. The first diffusion map clusters the data into three parts, the second of which is the area of interest. The other two sections are eliminated from the remaining calculations. In the second step, the remaining area is subjected to another diffusion map assessment and the internal layers are localized based on their textural similarities. The proposed method was tested on 23 datasets from two patient groups (glaucoma and normals). The mean unsigned border positioning errors (mean ± SD) was 8.52±3.13 and 7.56±2.95 μm for the 2D and 3D methods, respectively.

## 1. INTRODUCTION

Optical coherence tomography (OCT) is a powerful imaging modality used to image various aspects of biological tissues, such as structural information, blood flow, elastic parameters, change of polarization state, and molecular content [Huang et al. 1991]. OCT uses the principle of low coherence interferometry to generate two- or three-dimensional images of biological samples by obtaining high-resolution cross-sectional back-scattering profiles (Fig. 1(a)). In contrast to OCT technology development which has been a field of active research since 1991, OCT image segmentation has only been more fully explored during the last decade. Segmentation remains one of the most difficult and at the same time most commonly required steps in OCT image analysis. No typical segmentation method exists that can be expected to work equally well for all tasks [DeBuc 2011]. We may classify the OCT segmentation approaches into five distinct groups according to the image domain subjected to the segmentation algorithm. Methods applicable to 1) A-scan, 2) B-scan, 3) active contour approaches (frequently in 2-D), 4) analysis methods utilizing pattern recognition, and 5) segmentation methods using 3D graphs constructed from the 3D OCT volumetric images.

**A-Scan** methods were introduced by Hee [Hee et al. 1995]. Since only 2D OCT scanning devices were available at that time, the contribution from 3D context could not be investigated. The mentioned methods also suffered from long computation times [George et al. 2000], [Koozekanani et al. 2001] and had limited layer detection accuracy. Additional A-scan approaches have been introduced recently [Fabritius et al. 2000, Koprowski Wrobel 2009, Lu et al. 2010]. Fabritius [Fabritius et al. 2000] incorporated 3D intensity information and segmented the ILM and RPE directly from the OCT data without massive pre-processing in about 17-21 seconds per 3-D volume with errors smaller than 5 pixels in 99.7% of scans.

**B-Scan** methods allowed dealing with 2D noise by incorporating better denoising algorithms during the preprocessing step. However, the dependency on noise reduction requires complicated and time-consuming denoising methods such as anisotropic diffusion, which made these algorithms slow [Boyer et al. 2006, Baroni et al. 2007]. Additionally, the underlying intensity based methods and the relevant threshold selection were a problem that is case-dependent.

**Active contour** approaches for OCT image segmentation were first proposed by Cabrera Fernández [Fernández et al. 2004] and modified by Yazdanpanah [Yazdanpanah et al. 2009]. Time complexity and exact error reports are not available for the mentioned studies, which make comparison of these methods with other published methods difficult. Regardless, active contour algorithms surpass the performance of intensity based B-scan approaches, both in resistance to 2D noise and in accuracy.

**Pattern recognition** based approaches were presented in [Fuller et al. 2007, Mayer et al. 2008] and relied on a multi-resolution hierarchical support vector machine (SVM) or on fuzzy C-means clustering techniques. The former reported to have low ability in detection (6 pixels of line difference and 8% of thickness difference) and a high time complexity (2 minutes). The latter reported to have better results by 2 pixels of linear difference and 45 seconds of processing time. The later-introduced methods like graph-based approaches outperformed these techniques in both the accuracy and processing time.

**3D graph-based** methods seem so far to be suitable for the task in comparison to the above-mentioned approaches (despite the fact that newly developed A-Scan methods proved to be able to produce accurate and fast results). Their time requirements can be reduced to about 45 seconds per 3D volume (480×512×128 voxels) and they routinely achieve high accuracy with about 2.8μm (or less than 2 pixels) of layer-surface segmentation error. Such methods take advantage of newly developed 3D imaging systems, which provide better visualization and 3D rendering of segmentation results [Abràmoff et al. 2009, Lee et al. 2010, Quellec et al. 2010]. By design benefitting from contextual information represented in the analysis graph, these methods are robust to noise and do not require advanced noise reduction techniques in the preprocessing steps. While there is no theoretical limit on the number of layers that can be simultaneously segmented by these approaches, up to 10 layers are routinely identified in retinal OCT images.

TABLE I gives an overview of the mentioned approaches and compares the OCT systems, preprocessing methods, error ranges and computation times.

In this paper, we focus on novel spectral-geometric methods for graph based image segmentation and explore a two-step diffusion map approach for segmentation of OCT images. Our research demonstrates the ability of diffusion maps to segment gray-level images. To the best of our knowledge, this is the first report of such a method for OCT segmentation. "Diffusion maps" have a wide range of application in medical image segmentation [Andersson et al. 2008, Shen Meyer 2006, Wassermann et al. 2008, Neji Langs 2009]. However, it seems that investigators so-far employed images from high dimensionality modalities like Diffusion MRI, fMRI and/or spectral microscopic images. The OCT images analyzed here are simple single-channel gray level image datasets.

In this context, two important points may be noted: A: Every algorithm capable of dealing with high dimensionality is able to handle low dimensional (and even one-dimensional) datasets. B: In order to reduce the effects of unavoidable noise in OCT images and to get rid of very complicated and time-consuming noise reduction in the preprocessing step, more than one pixel (or voxel) can be grouped to represent nodes of our graph and three categories of textural features (statistics, co-occurrence matrix, run-length matrix) can be derived from each node to measure similarity between nodes. Such features can increase the dimensionality of the dataset and give advantage of dimensionality reduction properties of the diffusion maps.

TABLE I
A BRIEF LOOK AT DIFFERENT APPROACHES IN OCT SEGMENTATION. IT SHOULD BE NOTICED THAT THE NUMBER AS REPORTED CANNOT BE USED FOR DIRECT COMPARISON OF THE RELATIVE PERFORMANCES, SINCE DIFFERENT SETTINGS ARE UTILIZED IN EACH METHOD.

| Segmentation Approach | Papers | OCT systems | Preprocessing method | Error range | Computation time |
|---|---|---|---|---|---|
| A-scan | [Hee et al. 1995], [Huang et al. 1991], [George et al. 2000], [Koozekanani et al. 2001], [Gregori Knighton 2004], [Herzog et al. 2004], [Shahidi et al. 2005], [Ishikawa et al. 2005], [Srinivasan et al. 2008], [Fabritius et al. 2000], [Koprowski Wrobel 2009], [Lu et al. 2010], | TD-OCT (Humphrey 2000, Stratus, OCT 3 Carl-Zeiss Meditec) / SD OCT (Cirrus HD-OCT) | Low-pass filtering, 2D linear smoothing, median filter, non-linear anisotropic filter, Intensity signal based thresholding segmentation | 20-36 μm and around 5 pixels in recent papers | Not reported in older cases, but in recent ones like Fabritius [6], required about *17-21seconds* for each boundary using a PC With 2.4 GHz CPU, 2.93 GB RAM |

| | [Mayer et al. 2010] | | | | |
|---|---|---|---|---|---|
| B-scan | [Boyer et al. 2006], [Baroni et al. 2007], [Tan et al. 2008], [Bagci et al. 2008], [Kajić et al. 2010] | TD OCT (OCT 3000 Zees-Humphrey, OCT 2 Carl-Zeiss Meditec, Stratus) / SDOCT (RTVue100 OCT, Optovue, Freemont, CA) | 2D median filter, Gaussian smoothing filtering, bilateral filter | 4.2-5 μm | *9 seconds* on a Pentium, 1.8 GHz processor with 1 GB of RAM |
| Active contours | [Fernández et al. 2004], [Mishra et al. 2009], [Yazdanpanah et al. 2009], [Mujat et al. 2005] | TD OCT (Stratus OCT)/ experimental HR OCT (high speed) / experimental FD-OCT | Nonlinear anisotropic diffusion filter, adaptive vector-valued kernel function | Around 3 pixels | *5-84 seconds* using Pentium 4 CPU, 2.26 GHz |
| Artificial intelligence | [Fuller et al. 2007], [Mayer et al. 2008] | experimental 3D OCT, SD OCT (Spectralis) | SVM approach, 2D mean filter, directional filtering | Around 6 voxels | *45-120 seconds* on a 2GHz Pentium IV on a computer with 3GB of RAM (dual processor 3GHz Intel Xeon) |
| 2D or 3D graphs | [Garvin et al. 2008], [Abràmoff et al. 2009], [Lee et al. 2010], [Yang et al. 2010], [Quellec et al. 2010], [Chiu et al. 2010], [Chiu et al. 2012] | TD OCT (Stratus OCT) / SD OCT (Cirrus–OCT Topcon 3D OCT-1000) | 2D spectral reducing anisotropic diffusion filter, median filtering, wavelets | 2.8 - 6.1 μm | *45- 300 seconds* using a Windows XP workstation with a 3.2GHz Intel Xeon CPU/on a PC with Microsoft Windows XP Professional x64 edition, Intel core 2 Duo CPU at 3.00GHz, 4 GBRAM, *16 seconds* in fast segmentation mode |

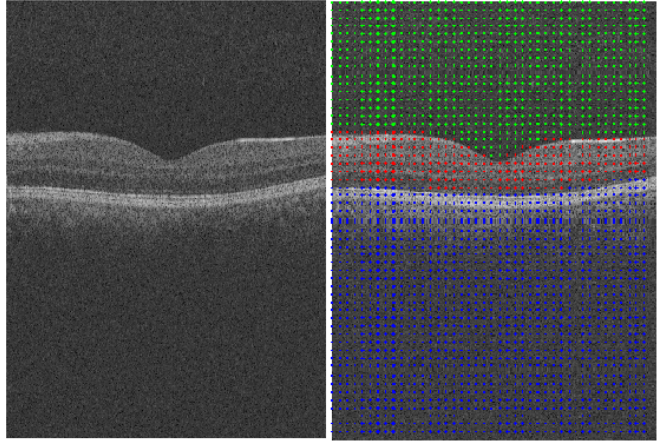

Fig.1. Left) Example OCT image. Right) Results of the first diffusion map shown on the right panel, note that the image is partitioned in 3 regions.

## 2. DIFFUSION MAPS

Diffusion maps [Coifman Lafon 2006] form a spectral embedding of a set X of n nodes, for which local geometries are defined by a kernel $k: X \times X \to R$. The kernel $k$ must satisfy $k(x,y) \geq 0$, and $k(x,y) = k(y,x)$. One may consider the kernel as an affinity between nodes which results in a graph (an edge between $x$ and $y$ carries the weights $k(x,y)$). The graph can also be defined as a reversible Markov chain by normalized graph Laplacian construction. We define

$s(x) = \sum_y k(x,y)$ (1)

and

$p_1(x,y) = \frac{k(x,y)}{s(x)}$ (2)

This new kernel is not symmetric, but it satisfies the requirements of being the probability of the transition from node $x$ to node $y$ in one time step, or a transition kernel of a Markov chain:

$\forall x, \sum_y p_1(x,y) = 1$ (3)

$P$ is the Markov matrix whose elements are $p_1(x,y)$ and the elements of its powers $P^\tau$ are the probability of the transition from node $x$ to node $y$ in $\tau$ time steps. The geometry defined by $P$ can be mapped to Euclidean geometry according to eigen-value decomposition of $P$.

Such a decomposition results in a sequence of eigen-values $\lambda_1, \lambda_2, \ldots$ and corresponding eigen-functions $\psi_1, \psi_2, \ldots$ that fulfill $P\psi_i = \lambda_i \psi_i$. The diffusion map after $\tau$ time steps $\Psi_\tau: X \to R^\omega$ embeds each node i = 1, …, n in the Markov chain into an $\omega$-dimensional Euclidean space. Any kind of clustering like k-means may be done in this new space.

$$i \to \Psi_\tau(i) = \begin{pmatrix} \lambda_1^\tau \psi_1(i) \\ \lambda_2^\tau \psi_2(i) \\ \vdots \\ \lambda_\omega^\tau \psi_\omega(i) \end{pmatrix} \qquad (4)$$

It is usually proposed to use a Gaussian kernel for the kernel $k(.,.)$, i.e. $k(x,y) = \exp\left(-\frac{d^2(x,y)}{2\sigma^2}\right)$, where $d$ is a distance over the set X and $\sigma$ is a scale factor.

## 2.1. COARSE GRAINING

This step identifies the best possible clustering of the data points. As described above, $P^\tau$ has elements which show the probability of the transition from node $x$ to node $y$ in $\tau$ time steps (the time $\tau$ of the diffusion plays the role of a scale parameter in the analysis; however, we don't want to go through other scales in this application and thus we chose τ=1). This parameter can be useful in upcoming research relating the diffusion map (Fourier on graphs) to diffusion wavelet (time-frequency approach on graphs), where scaled versions of our graphs are needed. The consequences of choosing other values than 1 for scale parameter $\tau$ is further elaborated in Section 3.2.

Below is a summary of coarse-graining approach in diffusion maps, proposed by Laofan [Lafon Lee 2006], provided here for completeness. The directly cited text is marked by quotes.

"If the graph is connected, we have:
$$\lim_{\tau \to +\infty} p^\tau(x,y) = \Phi_0(y), \qquad (5)$$
where $\Phi_0$ is the unique stationary distribution:
$$\Phi_0(x) = \frac{d(x)}{\sum_z d(z)} \qquad (6)$$
This distribution is proportional to the degree of x in the graph, which is a kind of measure for density of the points. It can also be shown that the following detailed balance condition is true:
$$\Phi_0(x) p^1(x,y) = \Phi_0(y) p^1(y,x). \qquad (7)$$"

"To relate distances to the spectral properties of the random walk, the $L^2$ metric between the conditional distributions can be defined and connect Markov random walk learning on graphs with data parameterization via eigen-maps."

Similar to [Nadler et al. 2006 (1)], Laofan in [Lafon Lee 2006] defined the "diffusion distance" $D_\tau$ between x and y as the weighted $L^2$ distance

$$D_\tau^2(x,z) = \|p^\tau(x,.) - p^\tau(z,.)\|^2_{1/\Phi_0} = \sum_{y \in \Omega} \frac{(p^\tau(x,y) - p^\tau(z,y))^2}{\Phi_0(y)} \qquad (8)$$

where the "weight" $\frac{1}{\Phi_0(y)}$ gives more penalty to domains with low density than those of high density.

Taking into account Eqs. (4) and (8), the diffusion distance can be approximated to relative precision $\delta$ using the first $q(\tau)$ non-trivial eigen-vectors and eigen-values according to:

$$D_\tau^2(x,z) \simeq \sum_{j=1}^{q(\tau)} \lambda_j^{2\tau}(\psi_j(x) - \psi_j(z))^2 = \|\Psi_\tau(x) - \Psi_\tau(z)\|^2 \qquad (9)$$

where $q(\tau)$ is the largest index j such that $|\lambda_j|^\tau > \delta|\lambda_1|^\tau$.

In order to retain the connectivity of graph points, resembling the intrinsic geometry, the coarse-grained graph $\tilde{G}$ with similar spectral properties should be constructed by grouping the nodes of the original graph. The weights between new nodes can also be obtained by averaging the weights between these new node sets. The important point is choosing the groups of nodes in a way that the quantization distortion is minimized.

"In order to construct the coarse-grained graph, a random partition $\{S_i\}_{1 \leq i \leq k}$ may be considered. Therefore, the members of each set $S_i$ are aggregated to form a node in a new graph ($\tilde{G}$) with k nodes. The weights of this graph can be calculated as:
$$\tilde{\omega}(S_i, S_j) = \sum_{x \in S_i} \sum_{y \in S_j} \Phi_0(x) p^\tau(x,y), \qquad (10)$$
where the sum accumulates all the transition probabilities between points $x \in S_i$ and $y \in S_j$ (Fig. 2)."

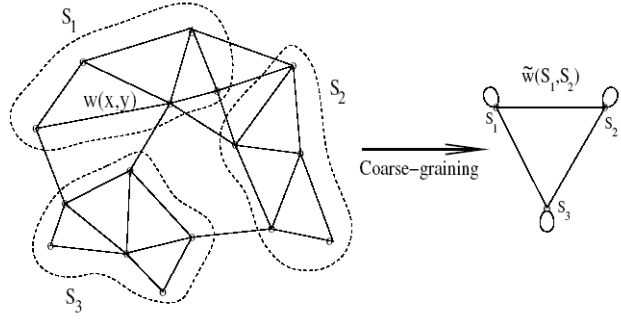

Fig. 2. Example of coarse-graining of a graph[Lafon Lee 2006].

"The new graph is symmetric and by setting
$$\tilde{\phi}_0(S_i) = \sum_{x \in S_i} \phi_0(x) \quad (11)$$
a reversible Markov chain can be defined on this graph:
$$\tilde{p}(S_i, S_j) = \frac{\tilde{\omega}(S_i, S_j)}{\sum_k \tilde{\omega}(S_i, S_j)} = \sum_{x \in S_i} \sum_{y \in S_j} \frac{\phi_0(x)}{\tilde{\phi}_0(S_i)} p_\tau(x, y) \quad (12)$$
where $\tilde{p}(.,.)$ are elements of matrix $\tilde{P}$, a $k \times k$ transition matrix of the newly developed coarse-grained matrix. The above definition can be similarly defined for $0 \leq l \leq n-1$.
$$\tilde{\phi}_l(S_i) = \sum_{x \in S_i} \phi_l(x) \quad (13)$$
According to duality condition, we may also define:
$$\tilde{\psi}(S_i) = \frac{\tilde{\phi}_l(S_i)}{\tilde{\phi}_0(S_i)} \quad (14)$$
The above vectors are expected to approximate the right and left eigen-vectors of $\tilde{P}$ and the choice of the partition $\{S_i\}$ is the most important factor in finding the best approximation."

"To have a measure of such an approximation, a new definition of the norm on coarse grained graph is also introduced:
$$\|\tilde{\phi}_l\|^2_{1/\tilde{\phi}_0} = \sum_i \frac{\tilde{\phi}_l^2(S_i)}{\tilde{\phi}_0(S_i)} \quad (15)$$
$$\|\tilde{\psi}_l\|^2_{\tilde{\phi}_0} = \sum_i \tilde{\psi}_l^2(S_i) \tilde{\phi}_0(S_i) \quad (16)$$
According to theses norms, a geometric centroid, or a representative point of each partition $S_i$ can be defined as:
$$c(S_i) = \sum_{x \in S_i} \frac{\phi_0(x)}{\tilde{\phi}_0(S_i)} \Psi_\tau(x)." \quad (17)$$

"For small values of $l$, $\tilde{\phi}_l$ and $\tilde{\psi}_l$ are approximate left and right eigen-vectors of $\tilde{P}$ with eigen-value $\lambda_l^\tau$. For $0 \leq l \leq n-1$:
$$\tilde{\phi}_l^T \tilde{P} = \lambda_l^\tau \tilde{\phi}_l^T + e_l \text{ and } \tilde{P}\tilde{\psi}_l = \lambda_l^\tau \tilde{\psi}_l + f_l \quad (18)$$
where:
$$\|e_l\|^2_{1/\tilde{\phi}_0} \leq 2D \text{ and } \|f_l\|^2_{\tilde{\phi}_0} \leq 2D \quad (19)$$
and:
$$D = \sum_i \sum_{x \in S_i} \phi_0(x) \|\Psi_\tau(x) - c(S_i)\|^2 \quad (20)$$
This means that if $|\lambda_l|^\tau \gg \sqrt{D}$, then $\tilde{\phi}_l$ and $\tilde{\psi}_l$ are approximate left and right eigen-vectors of $\tilde{P}$ with eigen-value $\lambda_l^\tau$."

"Namely, in order to maximize the quality of approximation, the following *distortion in diffusion space* should be minimized:
$$D = \sum_i \sum_{x \in S_i} \phi_0(x) \|\Psi_\tau(x) - c(S_i)\|^2 \approx E_i\{E_{x|i}\{\|\Psi_\tau(x) - c(S_i)\|^2 | X \in S_i\} \quad (21)$$
which can be written as a weighted sum of pairwise distances:
$$D = \frac{1}{2}\sum_i \tilde{\phi}_0(S_i) \sum_{z \in S_i} \sum_{x \in S_i} \frac{\phi_0(x)}{\tilde{\phi}_0(S_i)} \frac{\phi_0(z)}{\tilde{\phi}_0(S_i)} \|\Psi_\tau(x) - \Psi_\tau(z)\|^2 \quad (22)$$
or
$$D \approx \frac{1}{2} E_i\{E_{X,Z|i}\{\|\Psi_\tau(X) - \Psi_\tau(Z)\|^2 | X, Z \in S_i\}\}." \quad (23)$$

A connection to kernel k-means is defined in [Lafon Lee 2006] to define an algorithm for distortion minimization. A minimization algorithm based on solution of the problem of quantizing in diffusion space with k code-words based on mass distribution is a well-known minimization algorithm of the sample set $\Psi_\tau(\Omega)$.

**2.2. DETAILED DISCUSSION OF IMPLEMENTATION OF DIFFUSION MAPS**

Many papers have been presented to show the ability of diffusion maps in image segmentation [Andersson et al. 2008, Shen Meyer 2006, Wassermann et al. 2008, Neji Langs 2009] (Fig. 1(b)); however, none of them discussed the important details of this process.

**The first step** in diffusion maps is the construction of input data points (nodes of the graph) with desired dimension. For instance, in the case of clustering a point distribution, x and y coordinates of points represent the 2-dimensional input data; however, in gray level images, the intensity of points should be considered as the third dimension. There is no doubt that dimensionality can be increased in color images or images with particular patterns. Details describing selection of such features in a graph based method named "normalized cuts" can be found in [Shi Malik 2000]. Other possibilities of higher dimensionality in gray level images are also discussed later in this paper.

The important point in selection of data points is that it is not necessary to assign one pixel to one node; instead, we may aggregate a group of pixels/voxels together to form a node which may also be helpful in noise management. Details of selecting proper groups of pixels are presented in Section 3.

**The second step** is construction of distance functions (geometric and feature distances). A distance function is defined by matrices with a size of $N \times N$, where $N$ is the number of input points (nodes of the graph). To form the geometric distance, each element of the matrix is calculated as the Euclidean distance $\|X_{(i)} - X_{(j)}\|_2$ (or any other distance measure like Mahalanobis, Manhattan, etc.) of points; however, to construct the feature distance, the Euclidean distance of features $\|F_{(i)} - F_{(j)}\|_2$ is used. The range of each distance matrix is also calculated to show how wide is the distribution of the data set and to estimate the value of the scale factor ($\sigma$). Namely, if the distance matrix $\|F_{(i)} - F_{(j)}\|_2$ $\left(or\ \|X_{(i)} - X_{(j)}\|_2\right)$ range from one point to another, the scale factor $\sigma_{feature}$ (or $\sigma_{geo}$) is recommended to be 0.15 times this range.

**The third step** is construction of the kernel (weights of the graph). This kernel can be defined as:

$$k(i,j) = \exp\left(-\frac{\|F_{(i)} - F_{(j)}\|_2^2}{2\sigma_{feature}^2}\right) \cdot \begin{cases} \exp\left(-\frac{\|X_{(i)} - X_{(j)}\|_2^2}{2\sigma_{geo}^2}\right) & \text{if } \|X_{(i)} - X_{(j)}\|_2 < r \\ 0 & otherwise. \end{cases} \quad (24)$$

where $r$ determines the radius of the neighborhood that suppresses the weight of non-neighborhood nodes and consequently makes a sparse matrix.

The weight matrix should also be normalized:
$$p = D^{-1}k, \qquad (25)$$
where D is the diagonal matrix consisting of the row-sums of $k$. In this step, a threshold ($5 \times 10^{-6}$ in our case) can be selected to turn the low values to zero and make a more consistent sparse matrix named symmetric graph Laplacian or Markov matrix.

**The fourth step** is calculating the eigen-functions of the symmetric matrix. We use the LAPACK [Shi Malik 2000] routine for singular value decomposition and obtain the normalized right and left eigen-vectors of the Markov matrix. The normalization can be simply achieved with dividing the eigen-vectors to their first value. The right eigen-vectors are rescaled with eigen-values to calculate the diffusion coordinates in Eq. (4).

**The fifth step** is applying the k-means clustering on diffusion coordinates and iterating the algorithm for many times to select the clustering result for distortion minimization with coarse graining (Figs. 7 and 12).

**The sixth step** is recovering the input data points (nodes) corresponding to each of the clustered diffusion coordinates and replacing the graph partitioning with an image segmentation task.

Since the clustering algorithm is prone to produce very small groups (even single pixels as clusters), we may recreate the clustering with a higher number of k in cases that small grouping is not desirable.

### 3. IMPLEMENTING DIFFUSION MAPS ON GRAY-LEVEL IMAGES

In order to apply the diffusion maps to OCT images, we should first define the graph nodes on the image. The 2D data, on which we apply the algorithms, have size of 200×1024 and 512×650 pixels in our two evaluations. A complete description of the datasets is given in Sections 5.1 and 5.2.

#### 3.1. SPARSE GRAPH REPRESENTATION

In order to reduce the noise in OCT images and to get rid of the complicated and time consuming noise reduction in a preprocessing step[Shi Malik 2000] (like anisotropic diffusion proposed in many papers[Shi Malik 2000, Shi Malik 2000, Garvin et al. 2008]), we may join more than one pixel (or voxel) as the nodes of our graph and select three categories of textural features (Statistics, Co-occurrence Matrix, Run-Length Matrix) from each node to measure the similarity between the nodes. A vector is constructed containing the selected features of each node such as mean value (more sophisticated textural features are optional and not part of our implementation). Euclidean (L2) norm was used. To prevent uneven influence of features, feature values are normalized to have identical ranges. Selection of a group of pixels in order to represent each node of the graph, named sparse graph representation, may be done in many different ways. One of the easiest choices may be to define a square or rectangle of a pre-defined size and divide the whole image to non-overlapping or overlapping windows of the selected size. It should be noted that such a selection may significantly reduce the resolution, particularly on edges which may be located inside of these windows.

Another way to select multiple points as each node of the graph is selecting the super-pixels [Ren Malik 2003, Mori et al. 2004, Mori 2005]. The result of applying super-pixels on an OCT image is shown in Fig. 3.

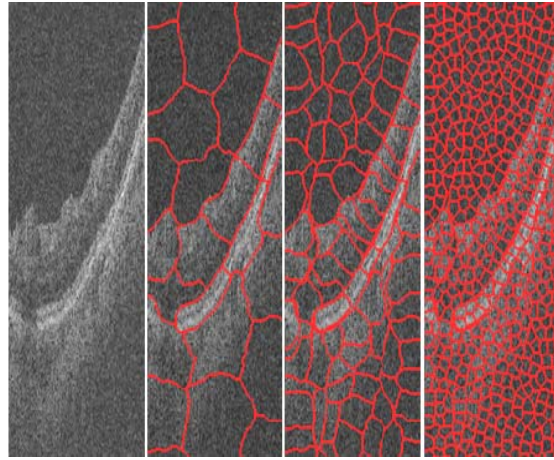

Fig.3. Result of applying super-pixel in 3 steps to an OCT image.

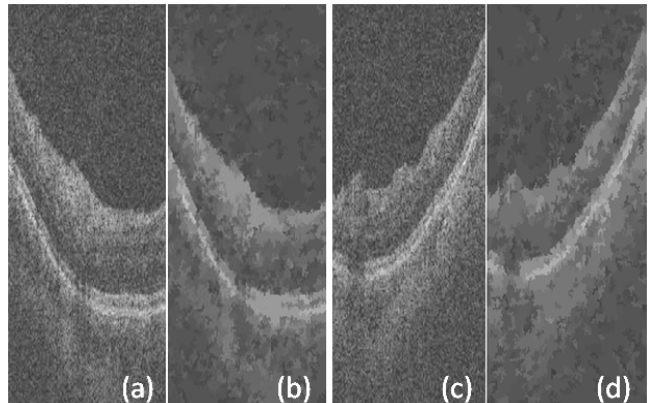

Fig. 4. Results of applying mean shift to an OCT image. (a, c) Original images. (b, d) Results of mean shift algorithm

It can be seen in Fig.3 that this method is able to select a group of pixels to represent a node in such a way that each group has the lowest possible amount of edges. This can simply lead to less reduction of resolution which occurred in the rectangular window method, described above. However, the very high time complexity of the super-pixel method (around 3 minutes for each image with a size of 650×512 pixels), even using mex files in C++, makes this method undesirable for this application.

Another method which may be useful for selection of groups of pixels as graph nodes is the mean shift algorithm [Paris Durand 2007]. This method combines the pixels with high degree of similarity and the combined pixels form the nodes of our graph. Fig.4 shows two examples of such results. However, this method reduces the essential details of OCT images and merges many desired layers which makes this procedure inappropriate for our images.

Comparing the proposed strategies (rectangular windowing, super-pixels and mean shift), we found the rectangular method more efficient from a computational and time complexity point of view. The most important drawback of the rectangular windowing algorithm is the decreased resolution which can be improved using a consequent converging step, like active contour searching for the highest gradient to locate the boundaries with an acceptable accuracy.

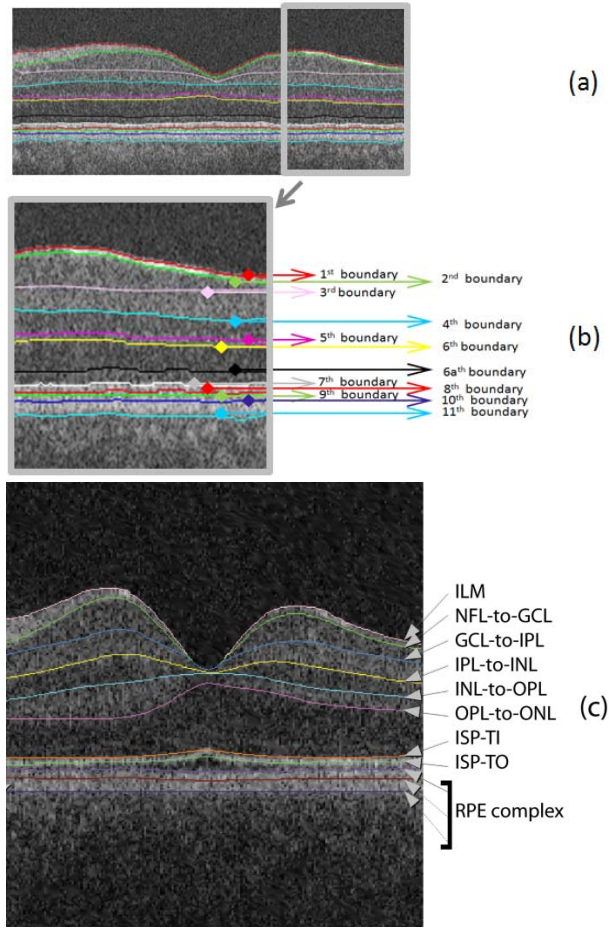

Fig.5. (a, b) Final segmentation on the original image. (c) Definition of eleven retinal surfaces (surfaces 1 – 11), ILM = internal limiting membrane, NFL = nerve fiber layer, GCL = ganglion cell layer, IPL = inner plexiform layer, INL = inner nuclear layer, OPL = outer plexiform layer, ONL = outer nuclear layer, ISP-TI = Inner segment of photoreceptors, transition to outer part of inner segment, ISP-TO = Inner segment of photoreceptors, start of transition to outer segment, RPE = retinal pigment epithelium.

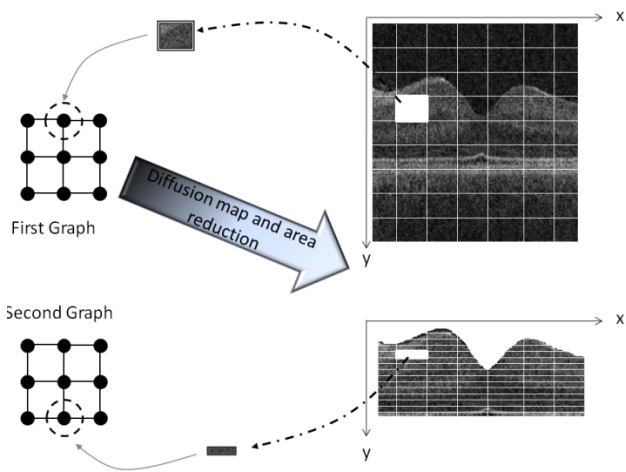

Fig.6. Rectangular windowing on 2D OCT images (in two distinct steps).

### 3.2. RECTANGULAR WINDOWING ON OCT IMAGES

In order to apply the diffusion maps to an OCT image, graph nodes must be associated with the image pixels/voxels. We employ the diffusion map in 2 sequential steps, the first of which segments 6 boundaries simultaneously, i.e., the 1st and 7th to 11th boundaries. The second step identifies the inner boundaries, i.e., 2nd to 6th boundaries and the new boundary (which we call 6a) as described in Fig. 5. For implementing the first step, we select $10 \times 10$ pixel boxes as graph nodes (Fig.6) and the kernel is defined as:

$$k(x,y) = \exp-\left(\left(\frac{d^2(x,y)}{2\sigma_{geo}^2}\right) + \left(\frac{d^2(g(x),g(y))}{2\sigma_{gray}^2}\right)\right) \quad (26)$$

where $x, y$ indicate the centroids of selected $10 \times 10$ boxes, $g(.)$ is the mean gray level of each box, and $\sigma_{geo}$ and $\sigma_{gray}$ point out the scale factor (calculated as 0.15 times the range of $d(x,y)$ and $d(g(x),g(y))$, respectively).

According to the rules of diffusion maps described in Section 2, we formed the $\omega = 3$ dimensional Euclidean space:

$$i \to \Psi_\tau(i) = \begin{pmatrix} \lambda_1^\tau \Psi_1(i) \\ \lambda_1^\tau \Psi_1(i) \\ \lambda_3^\tau \Psi_3(i) \end{pmatrix} \quad (27)$$

Subsequently, k-means clustering with k=3 followed by coarse graining is applied to the Euclidean space constructed by eigen-functions (Figs.1(b) and 7). Since the results of k-means clustering are prone to getting different clustering results in every single run (as a result of randomization of the seed, an intrinsic drawback of this method in arbitrary selection of start points), best possible clustering of the data points is achieved by minimizing the quantization distortion using the coarse graining approach.

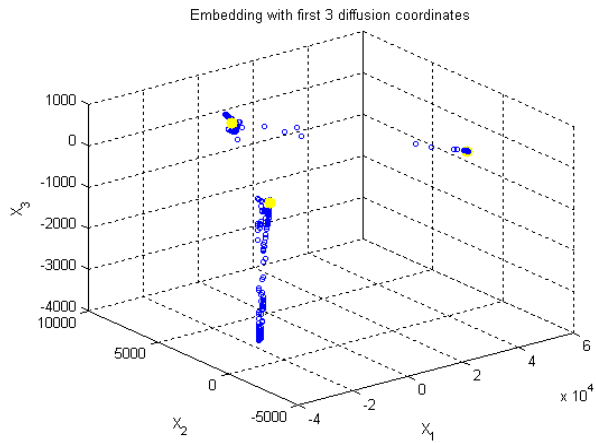

Fig.7. Applying k means clustering on diffusion coordinates of the first diffusion map, the cluster centroids are shown in yellow and 3 axes are 3 most important eigen functions.

The edge points of the upper and lower clusters are extracted (Fig.8, top-left) by locating the smallest and the highest values of vertical range in each column. Afterwards, they are connected using a simple linear interpolation (Fig.8, top-right). It is obvious that the resulting curves are not located exactly on the particular position of edges, due to the size of selected boxes in the first step of diffusion maps. Therefore, the extracted edges are moved to the lowest vertical gradient (Fig.8, bottom-right and Fig.9) in a vertical search area of 10 pixels above and 10 pixels below. The results are then enhanced based on applying the following operators on the edges to remove the outliers and to produce a smooth curve: cubic spline smoothing, local regression using weighted linear least squares, and 2nd degree polynomial models (Fig.8, bottom-right).

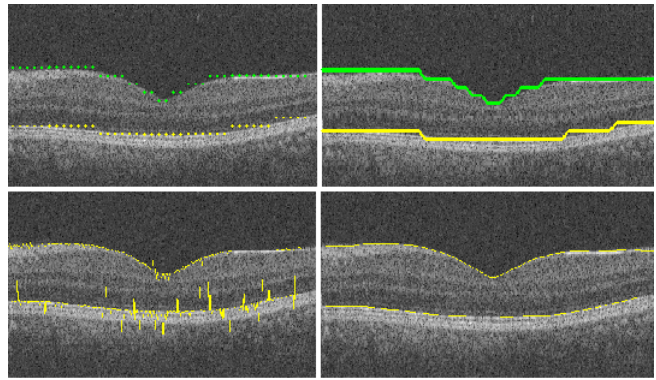

Fig.8. Top-left) extracted edge points of the upper and lower clusters. Top-right) after interpolation and cubic spline smoothing. Bottom-left) after moving to highest gradient. Bottom-right) after local regression using weighted linear least squares, and $2^{nd}$ degree polynomial models.

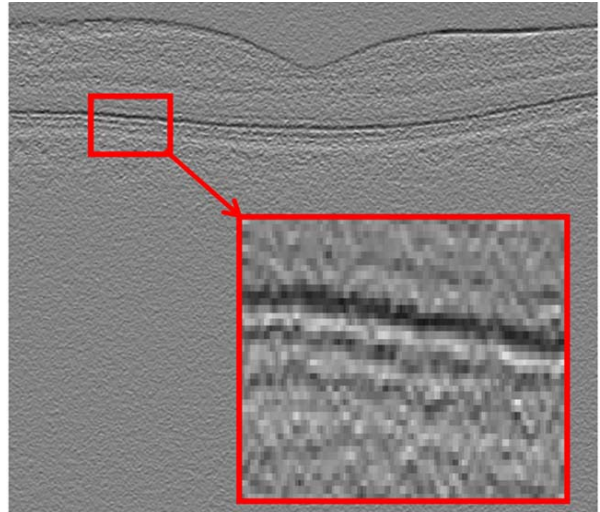

Fig.9. Vertical gradient.

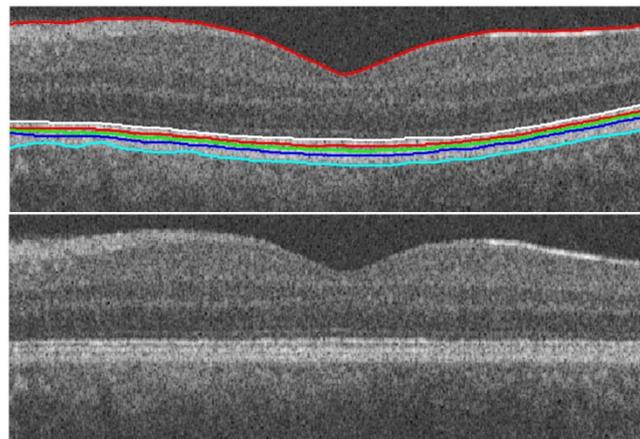

Fig.10. Top) $1^{st}$ and $7^{th}$ to $11^{th}$ boundaries. Bottom) Linearization.

$1^{st}$ and $7^{th}$ boundaries are obtained in this step and the $8^{th} - 11^{th}$ boundaries are detected by looking for the highest and lowest (alternatively) vertical gradients in a vertical search area of 10 pixels below (Fig.10, top).

The unwanted drift of the OCT images is then removed according to the $10^{th}$ boundary to change each column of image in order to produce a linear layer in the place of the $10^{th}$ border (Fig.10, bottom).

As described in Section 2.1, time of the diffusion plays the role of a scale parameter in the analysis and using higher powers of the kernel ($\tau > 1$) are similar to applying dilation on our graph or calculating the $\tau$-step connections between the nodes. We found that meaningful segmentation results when using single-step connections ($\tau = 1$), while other values of $\tau$ resulted in dispersed clusters and undesired divisions. Fig.11 shows three samples of applying multi-step connections on our first diffusion map.

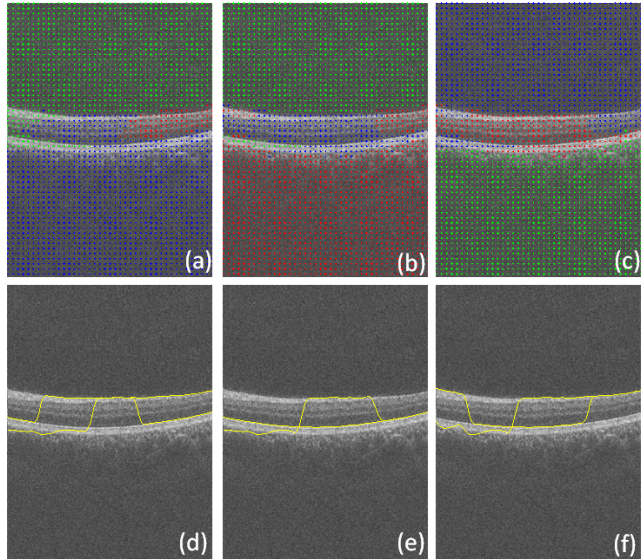

Fig. 11. Three samples of applying multi-step connections on our first diffusion map. (a-c) Clusters for τ = 2, 3, 4, respectively. (d-f) Estimated 1$^{st}$ and 7$^{th}$ boundaries.

Another parameter, selection of which should be discussed, is the scale factor ($\sigma$). As mentioned in section 2.2, the range of each distance matrix should be found and the scale factors are recommended to be calculated as 0.15 times this range. The choice of the parameter $\sigma$ is important for computation of the weights and is a data-dependent parameter [Bah 2008]. Small values of $\sigma$ will yield almost-zero entries for $k(x, y)$, while large values will lead to close-to-one values (desirable values lie between these two extremes). Based on this idea, Singer et al. [Singer 2007] proposed a scheme that was adopted in this work as follows:

1. Construct a sizeable $\sigma$-dependent weight matrix $k = k(\sigma)$ for several values of $\sigma$.
2. Compute $L(\sigma) = \sum_x \sum_y k_{xy}(\sigma)$.
3. Plot $L(\sigma)$ using a logarithmic plot. This plot will have two asymptotes when $\sigma \to 0$ and $\sigma \to \infty$.
4. Choose $\sigma$, for which the logarithmic plot of $L(\sigma)$ appears linear.

We employed the above algorithm in our application and found that selecting $\sigma$ equal to 0.15 times the range of distance matrix falls in the linear part of $L(\sigma)$ and thus there is no need to compute $L(\sigma)$ separately for each input dataset.

In the next step, as demonstrated in Fig.6, the search area is narrowed to pixels located between the 1$^{st}$ and 7$^{th}$ boundaries and the pixel boxes representing the graph nodes are selected as very thin horizontal rectangles ($2 \times 20\ pixels$). This selection is according to the structure of OCT images after linearization (Fig.10, bottom). The kernel is selected similar to Eq. (26) and the k-means clustering is applied with k=6 (Fig.12). It should be mentioned that the domain of applying the second diffusion map is decreased to the area located between 1$^{st}$ and 7$^{th}$ boundary and the rest of the image is eliminated. Therefore, the computation time decreases considerably and the segmentation becomes focused on a region of interest.

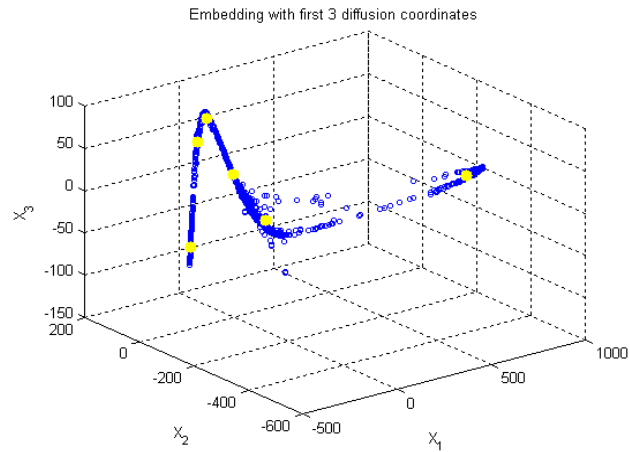

Fig.12. Applying the k means clustering on diffusion coordinates of the second diffusion map, the cluster centroids are shown in yellow and 3 axes are 3 most important eigen functions.
.

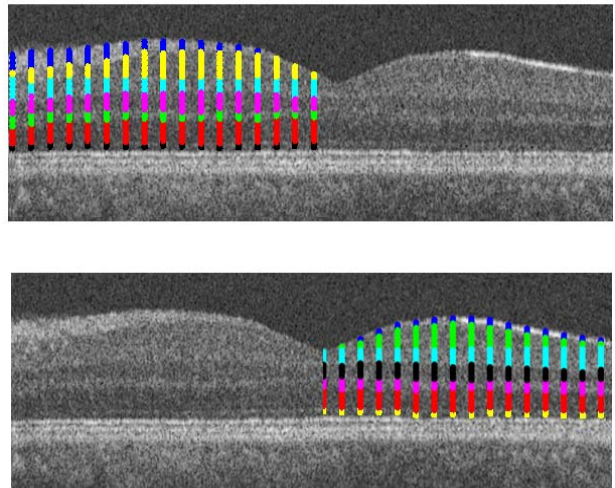

Fig.13. The results of the second diffusion map on the right and left parts of the image.

The image area was divided in two (right and left) parts to increase the accuracy of the method (Fig. 13). It should be mentioned that in the case of overall assessment of the images (without breaking to right and left parts), the algorithm couldn't find proper subgroups. For instance, the first layer in the right and left parts of Fig.5(a) has no connections and couldn't be merged to form a good cluster after employing the diffusion map algorithm.
The edge points of clusters were extracted and the points were connected to make a smooth curve (Fig.14).

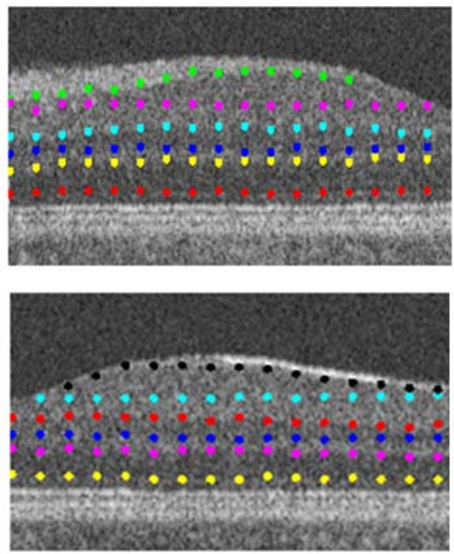

Fig.14. The smoothed curves on two sides.

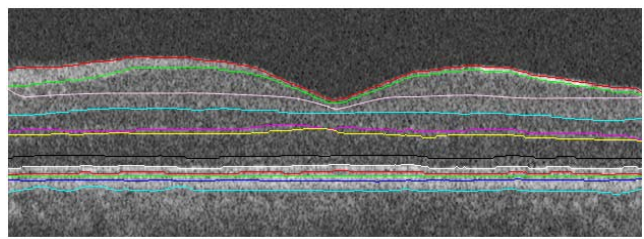

Fig.15. Final segmentation on the aligned image.

In the final step, the curves are connected together to form the final segmentation (Figs.15 and 5).
One may also use consecutive 2D algorithms applied to each layer of 3D OCT and use the combination of these boundaries to form the 3D surfaces. This idea was tested on our images; however, its time complexity made it impractical. Therefore, a new 3D algorithm was proposed to work on original 3D volumes (Section 4).

### 3.3. COMPUTATION OF THE NUMBER OF CLUSTERS

Accurate selection of the number of clusters in k-means clustering is important and can be done in at least two ways. The first method is described in [Nadler et al. 2006 (2)] and is based on detection of an "elbow" in the eigen-value plots. Briefly, if the slope of the eigen-value plot changes noticeably at the eigen-vector $\lambda_i$, the number of clusters should be i+1. The second way is based on the re-ordering of the rows and columns in the affinity matrix following the second eigen-vector, as proved in [Nadler et al. 2006 (2)], which shows the block structure of the matrix as squared blocks along the main diagonal. The number of clusters corresponds to the number of blocks.

The "elbow" algorithm in the first step of diffusion maps on 2D OCT data yields k=3 as the correct number of clusters (and was used in our algorithms). Fig.16(a) shows the eigen-value plot and the plot of $\frac{\lambda}{\lambda-1}$ to magnify the change of slope. We may similarly find that k=6 is, in general, the best cluster number for the second step of the diffusion map algorithm for 2D OCT data, which fully agrees with our selected clustering number (Fig.16(b)).

The most influential outcome of this method is its ability to determine the proper number of layers in an OCT image, which may vary due to different retinal disorders in the patient, or even as a consequence of using different OCT imaging systems. Fig. 17 shows how our method can be applied to patient data with merged or missing layers due to disease or limitations of the imaging system. The reported approach is capable of determining the desired number of clusters, a problem that could only be solved manually in the previously reported methods.

When determining the proper number of clusters, two consequences may be expected. If the chosen number is lower than desired, layers will merge; otherwise, if we choose a greater number, over-segmentation will occur. Fig. 18 shows examples of over-segmentation and merging in the first and second diffusion maps. It is obvious that turquoise colors in Fig. 18(a, d, e) represent the over-segmented areas and Fig. 18(b, f, g) demonstrates region merging.

More clarification of this approach and a step by step algorithm for automatic segmentation of OCT datasets can be found in section 5.3. It should also be mentioned that the outer retinal layers (numbered 7-11) are segmented using standard edge gradient consideration, and therefore these layers are not affected by choice of k in the second step of the algorithm.

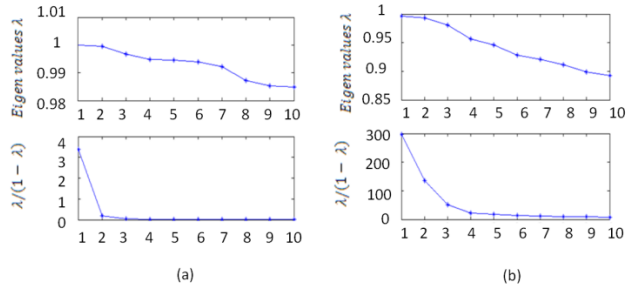

Fig.16. (a) Eigen-value plot in first step of diffusion maps on 2D OCT data. (b) Eigen-value plot in second step of diffusion maps on 2D OCT data.

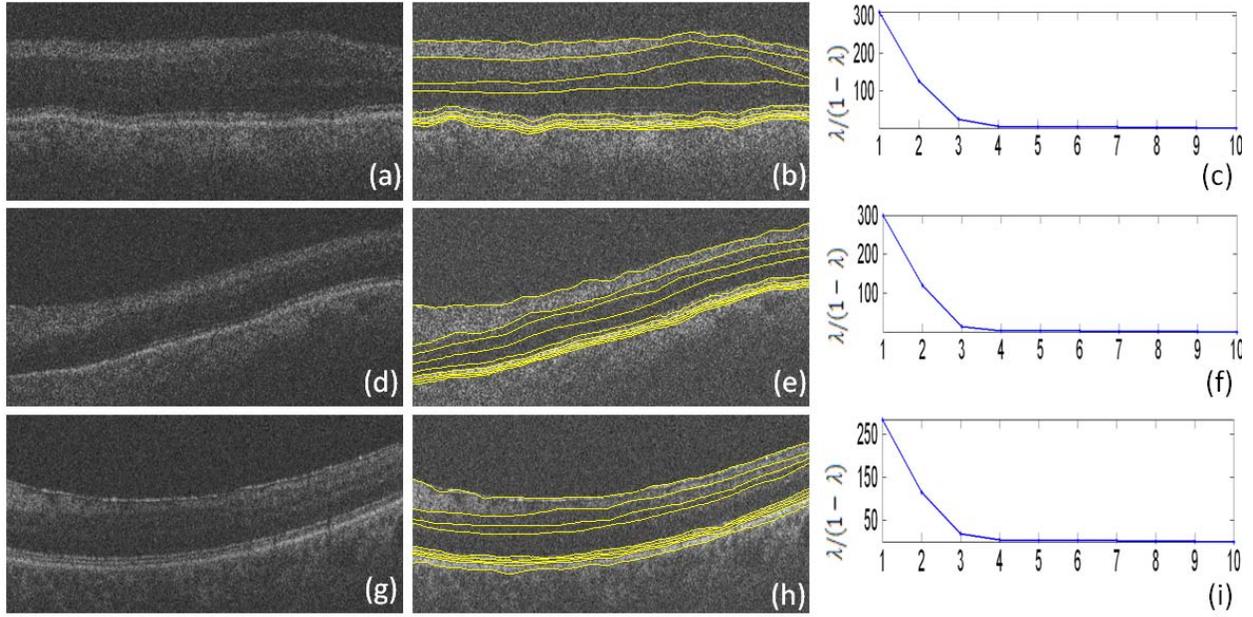

Fig. 17. Application of the proposed method on merged retinal layers (due to pathological appearance or imaging imperfections). (a, d, g) Original images. (b, e, h) Our segmentation method automatically identified the correct number of clusters to segment 8 layers (9 surfaces) compared to the usual number of layers. (c, f, i) Eigen-value plots in the second steps of the diffusion map algorithm.

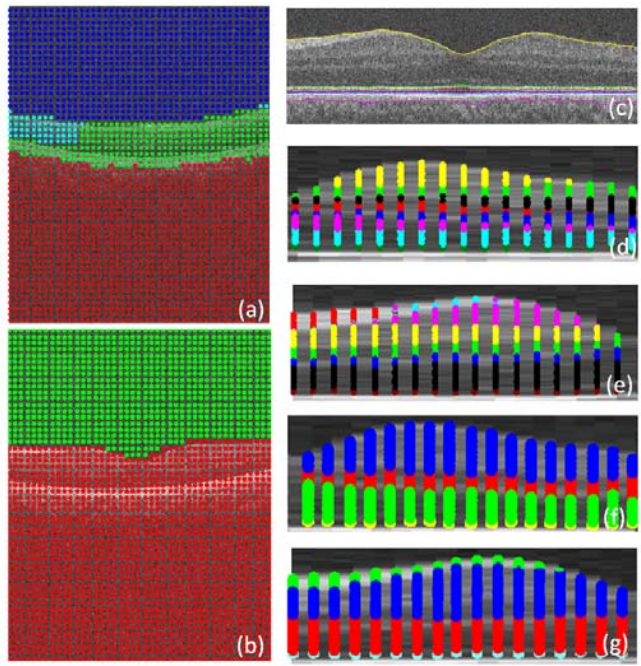

Fig. 18. Selecting an incorrect number of clusters. (a) First diffusion map. Oversegmentation with k=4. (b) First diffusion map, merging with k=2. (c) Original image. (d,e) Second diffusion map, oversegmentation with k=8. (f,g) Second diffusion map, merging with k=4.

## 4. IMPLEMENTING DIFFUSION MAPS ON 3D OCT DATA

To implement the diffusion maps on 3D OCT data, each node of the graph is allocated to a cubic portion of the whole 3D data. The 3D data, on which we apply the algorithms, has size of 200×200×1024 and 512×650×128 pixels in our two evaluations. A complete description of datasets is given in Sections 5.1 and 5.2.

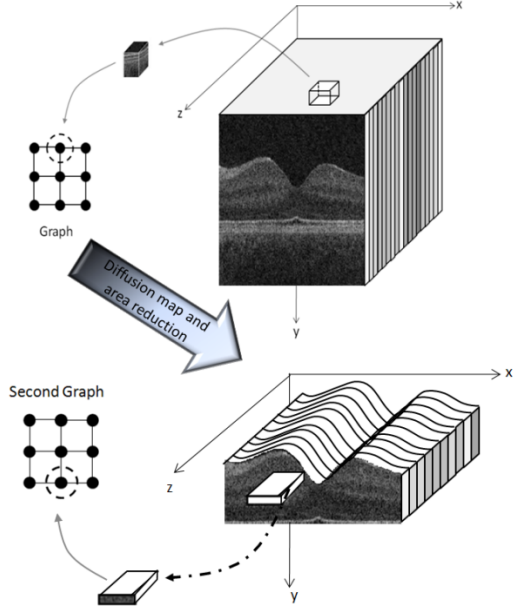
Fig.19. Construction of graph nodes from 3D OCT.

Similar to the two-dimensional approach described in Section 3, we employ the diffusion map in 2 successive stages (Fig. 19). In the first stage, 6 boundaries in 3D (the 1$^{st}$, and 7$^{th}$ to 11$^{th}$ boundaries, see Fig.5) are extracted concurrently. The second step is to identify the inner layers in 3D, i.e., the 2$^{nd}$ to 6$^{th}$ boundaries and the new boundary (labeled 6a). In the first step, $10 \times 10 \times 10$ pixel cubes are attached to form the graph nodes and the kernel is defined as:

$$k(x,y,z) = \exp\left(-\frac{d^2(x,y,z)}{2\sigma_{geo}^2}\right) \cdot \exp\left(-\frac{d^2(g(x),g(y),g(z))}{2\sigma_{gray}^2}\right) \quad (28)$$

where $x, y, z$ indicate the centroids of selected $10 \times 10 \times 10$ boxes, $g(.)$ is the mean gray level of each box, and $\sigma_{geo}$ and $\sigma_{gray}$ point out the scale factor (calculated as 0.15 times the range of $d(x,y,z)$ and $d(g(x),g(y),g(z))$, respectively). Similar to what is described in Sections 2 and 3, a three dimensional Euclidean space is constructed according to Eq. (27) and the k-means clustering with k=3 followed by coarse graining is applied to the Euclidean space constructed by eigen-functions (Fig.20).

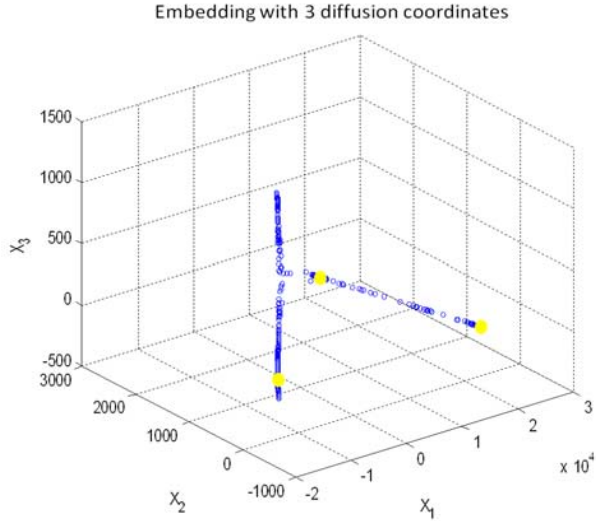

Fig.20. Applying the k means clustering on diffusion coordinates of the first diffusion map in 3D, the cluster centroids are shown in yellow and 3 axes are 3 most important eigen functions.

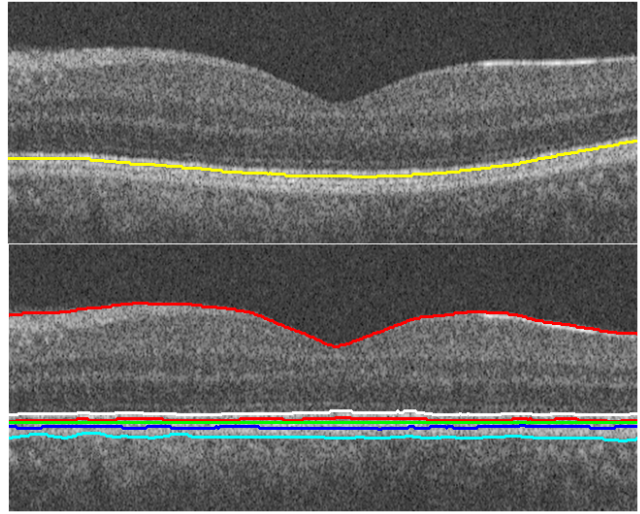

Fig. 21. Boundaries (like 1st and 7th) were detected only in selected slices, which coincided with the borders of the determined cubes. The boundaries in all remaining slices were determined using interpolation. Top) Border detection using interpolation. Bottom) Image alignment in 3D.

Cubic spline smoothing, local regression using weighted linear least squares, and 2nd degree polynomial models are sequential steps applied to the edge points of the upper and lower clusters in 3D. In order to move the points to the lowest vertical gradient, one may consider the 3D volume as a combination of consecutive 2D slices and search the vertical area in each slice; however, to overcome the time complexity, the gradient search was only applied in selected slices which were located in borders of the determined cubes with a size of $10 \times 10 \times 10$. The boundaries of remaining slices were then determined using simple interpolation (Fig.21, top). In the next step, the whole 3D volume was subjected to a 3D drift removing algorithm according to 10th boundary (Fig.21, bottom).

The following step is eliminating the area located above the 1st boundary and below the 7th boundary (Fig.22), choosing $15 \times 1 \times 15$ regions representing the graph nodes with a kernel (26), and applying the k-means clustering with k=6 (Fig. 23). The selected blocks are thin in the $y$ direction to provide the best possible resolution in detection of internal layers.

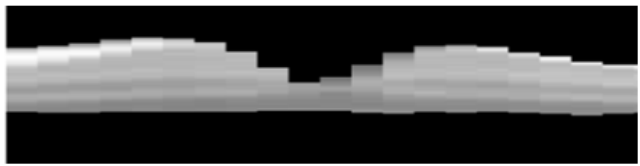

Fig. 22. Selecting the region of interest and choosing $15 \times 1 \times 15$ regions representing the graph nodes (the result is shown on one sample slice).

In the final step, 3D boundaries form the final segmentation can be shown (Figs.24 and 25).

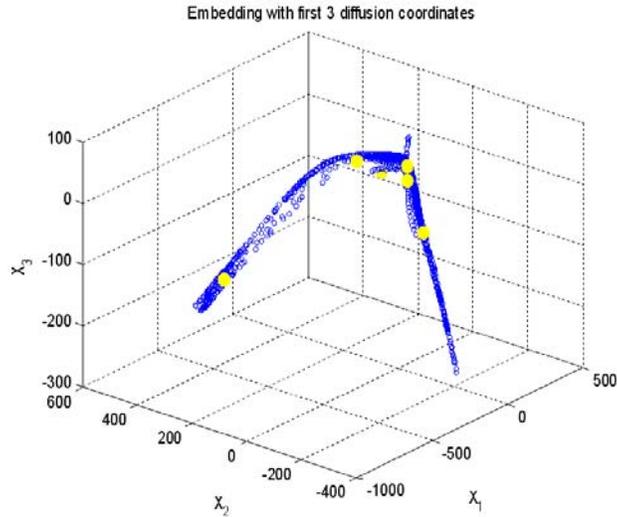

Fig.23. Applying the k means clustering on diffusion coordinates of the second diffusion map in 3D, the cluster centroids are shown in yellow and 3 axes are 3 most important eigen functions.

## 5. EXPERIMENTS AND RESULTS

The application of diffusion maps to segmentation of OCT images is presented in 2D and 3D. To evaluate the capability of this method, 23 datasets from two patient groups were analyzed. The first group consisted of 10 OCT images from 10 patients diagnosed with glaucoma [Antony 2010]. The independent standard resulted from averaging tracings from two expert observers and performance assessment results are given in Section 5.1. The second group (the Isfahan dataset) was obtained from 13 normal eyes from 13 subjects and the validation was based on averaged tracings by two expert observers (Section 5.2).

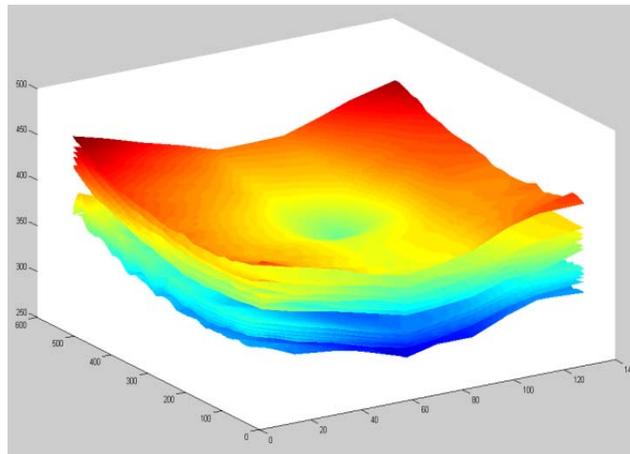

Fig. 24. Final segmentation on 3D OCT.

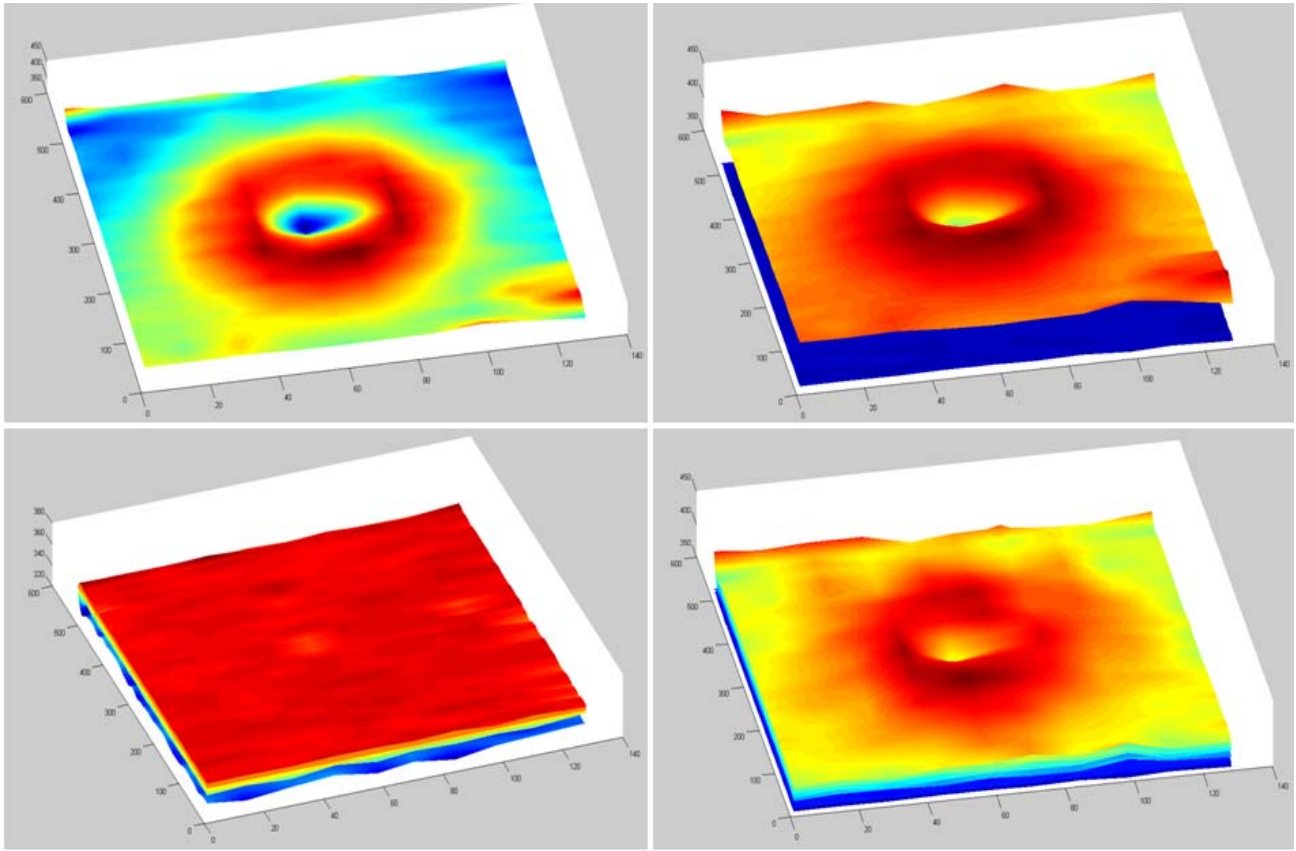

Fig. 25. Up-left) Surface 1. Up-right) 1st and 7th surfaces. Bottom-left) Surfaces 8 to11. Bottom-right) Surfaces 3 to7. All results have been produced after alignment based on the hyper-reflective complex (HRC) layer.

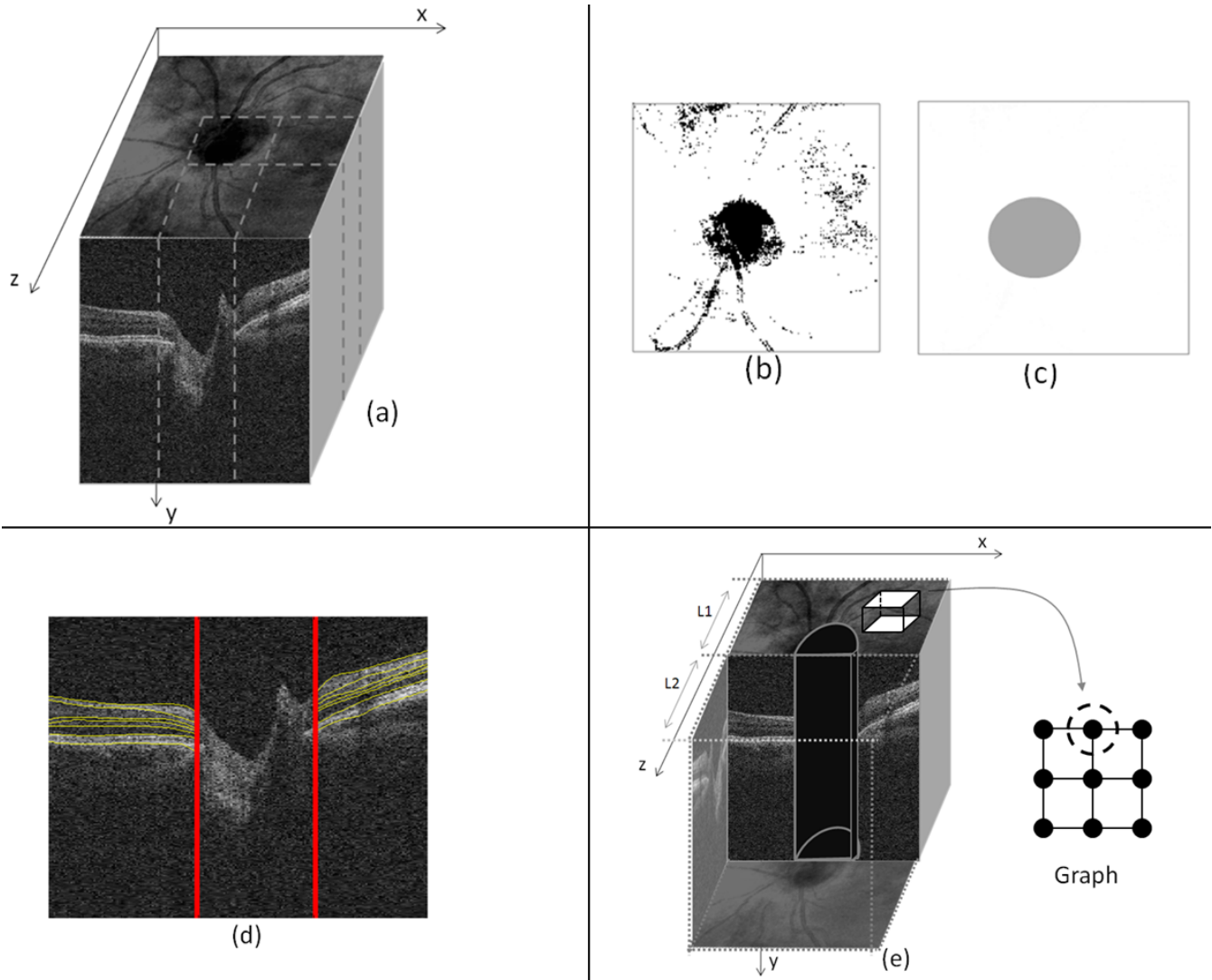

Fig. 26. Segmentation of images of the optic nerve head. (a) 3D data. (b) Projection image in x-z direction. (c) ONH mask made according to the projected image. (d) 2D segmentation of ONH images neglecting nerve canal area. (e) Construction of 3D graph while eliminating the central cylindrical volume.

To give additional insight into splitting the images into right and left parts, we may consider the four following states: For the 2D approach and in images of macula, we separate the central one third of the image and look for a possible minimum of the first boundary. If the minimum is located in the middle, we select it as the point from which the image should be split; otherwise, the minimum is on one of the two remaining thirds, which shows a gradual change of the boundary (e.g., Fig.28(a)) and the image can be split from a point in the middle of the x-axis. In practice, we don't need such a separation since the discontinuity for which we are planning the right and left splitting doesn't occur anymore.

For the 3D approach and in images of macula, no splitting is needed.

For the 2D ONH approach, an estimate of the ONH location can be obtained when making a projection image (mean intensity along the y-axis) from a number of slices (Fig. 26 (a, b, c)). According to the position of the ONH, the nerve head can be omitted from the x-y images and the segmentation is performed on two-sided regions (Fig. 26 (d)). For the alignment purposes, the curves are interpolated in the middle area and the images are aligned according to such interpolated curves.

For the 3D ONH approach, the ONH location can be estimated in the same way as described above. When constructing the graph, the nodes located within a cylinder with an elliptical horizontal cut can be disregarded (Fig. 26 (e)). After computing the first diffusion map, the produced surfaces are nonexistent in the middle part and can be interpolated for alignment purposes. The same applies to the other surfaces resulting from the second diffusion map but no interpolation of the neural canal region is needed (as mentioned in Section 5.1).

## 5.1. EXPERIMENTAL RESULTS ON THE GLAUCOMATOUS TEST SET

In order to have a comparison with results of Lee [Lee et al. 2010] and Antony [Lee et al. 2010], we applied our algorithm to a 2D-labeled glaucoma OCT dataset (Cirrus, Zeiss Meditec) [Antony 2010]. Two example results are shown in Fig.27. Since the independent standard available for the datasets contained a subset of 7 retinal surfaces, the proposed algorithm was only tested on these 7 surfaces.

The testing set for the validation of the segmentation algorithm consisted of 10 datasets from 10 patients diagnosed with glaucoma. Each dataset had x,y,z dimensions of $6 \times 2 \times 6$ mm$^3$, $200 \times 1024 \times 200$ voxels sized $30 \times 1.95 \times 30$ μm$^3$ (Fig.19.). The algorithm was tested against the manual tracings of two independent observers on 10 slices selected randomly (from 10 sections of 20 slices) in each dataset and the 7 retinal layers are traced [Antony 2010], Fig. 27.

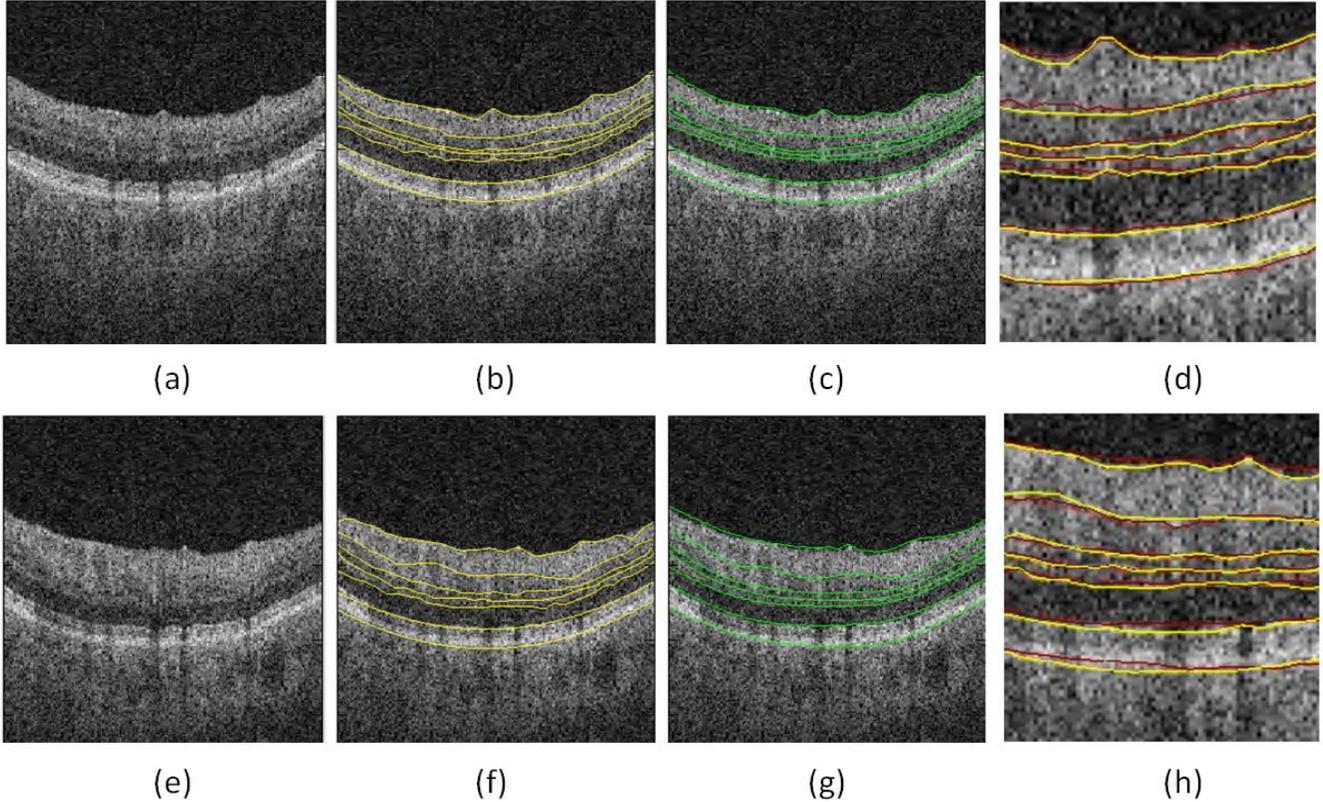

Fig. 27. Two example results obtained on the glaucoma test set. (a,e) Composite image. (b,f) Composite image with computer-segmented borders. (c,g) Composite image showing independent standard. (d,h) Comparison of computer-segmentation (yellow) and independent standard (red).

TABLE II
SUMMARY OF MEAN UNSIGNED BORDER POSITIONING ERRORS (MEAN ± SD) IN MICROMETERS

| Border | Avg. Obs. vs. Our Alg. | Avg. Obs. vs. Alg.in[Lee et al. 2010] | Avg. Obs. vs. Alg. in[Antony 2010] | Obs. 1 vs. Obs. 2 |
|---|---|---|---|---|
| 1 | 4.11±1.17 | 6.73±3.95 | 4.90 ± 1.54 | 4.90 ± 1.37 |
| 2 | 8.60±4.11 | 14.64±4.26 | 14.43 ± 5.63 | 12.79 ± 3.36 |
| 4 | 9.19±3.13 | 11.52±3.42 | 10.96 ± 4.06 | 13.74 ± 2.04 |
| 5 | 6.45±2.74 | 10.50±3.33 | 10.46 ± 2.79 | 9.28 ± 3.00 |
| 6 | 9.19±2.75 | 8.87±2.55 | 10.73 ± 2.78 | 7.67 ± 1.69 |
| 7 | 2.35±1.56 | 3.87±0.80 | 3.87 ± 1.32 | 4.69 ± 1.26 |
| 11 | 4.11±0.59 | 6.72±1.44 | 7.24 ± 1.74 | 6.58 ± 1.53 |
| overall | 6.32±2.34 | 8.98±3.58 | 8.94 ± 3.76 | 8.52 ± 3.61 |

For validation, the mean signed and unsigned border positioning errors for each border were computed and are presented in TABLES II and III for each boundary, differences were not computed in the neural canal. The algorithm is modified to trace only 7 surfaces out of 12 boundaries shown in Fig.5. To show statistically significant improvement of the proposed method over the algorithms reported in [Lee et al. 2010] and [Antony 2010], TABLE III shows the obtained p-values. Except for comparison of our method to Alg. [Lee et al. 2010] in the 6$^{th}$ boundary, our results show a significant improvement over the mentioned methods.

Furthermore, the errors of mean layer thicknesses determined as the distances from the first to other surfaces are computed and shown in TABLE V.

TABLE III
IMPROVEMENT OF THE PROPOSED METHOD COMPARED WITH ALGORITHMS [Lee et al. 2010] AND [Antony 2010]

| Border | p value Our Alg. vs. Alg. in [Lee et al. 2010] | p value Our Alg. vs. Alg. in [Antony 2010] |
|---|---|---|
| 1 | <<0.001 | <0.001 |
| 2 | <<0.001 | <<0.001 |
| 4 | <0.001 | <0.001 |
| 5 | <<0.001 | <<0.001 |
| 6 | 0.8022 | <0.001 |
| 7 | <<0.001 | <<0.001 |
| 11 | <<0.001 | <<0.001 |
| overall | <<0.001 | <<0.001 |

TABLE IV
MEAN SIGNED BORDER POSITIONING ERRORS (MEAN ± SD) IN MICROMETERS

| Border | Avg. Obs. vs. Alg. | Avg. Obs. vs. [Lee et al. 2010] |
|---|---|---|
| 1 | 1.56±1.04 | 2.55±1.19 |
| 2 | -7.56±4.36 | -8.33±7.83 |
| 4 | 7.04±4.13 | -9.19±3.67 |
| 5 | -5.28±2.64 | -5.44±4.10 |
| 6 | -6.84±2.19 | -6.63±2.96 |
| 7 | 1.17±0.92 | 1.54±1.09 |
| 11 | -1.76±1.49 | -1.86±3.01 |
| overall | -1.69±2.52 | -3.91±4.71 |

TABLE V
MEAN ABSOLUTE THICKNESS DIFFERENCES (MEAN ± SD) IN MICROMETERS

| Border | Avg. Obs. vs. Alg. |
|---|---|
| 1-2 | 6.1± 3.8 |
| 1-4 | 3.7± 2.2 |
| 1-5 | 1.9± 1.0 |
| 1-6 | 6.3± 3.2 |
| 1-7 | 2.0± 1.3 |
| 1-11 | 3.8± 2.7 |

For validation, the mean signed and unsigned border positioning errors for each border were computed and are presented in TABLES II and III for each boundary, differences were not computed in the neural canal. The algorithm is modified to trace only 7 surfaces out of 12 boundaries shown in Fig.5.To show statistically significant improvement of the proposed method over the algorithms reported in [Lee et al. 2010] and [Antony 2010], TABLE III shows the obtained p-values. Except for comparison of our method to Alg. [Lee et al. 2010] in the 6$^{th}$ boundary, our results show a significant improvement over the mentioned methods.

Furthermore, the errors of mean layer thicknesses determined as the distances from the first to other surfaces are computed and shown in TABLE V.

## 5.2. EXPERIMENTAL RESULTS ON THE ISFAHANTEST SET

The proposed method was tested on thirteen3D macular SD-OCT images obtained from eyes without pathologies using Topcon 3D OCT-1000 imaging system in Ophthalmology Dept., Feiz Hospital, Isfahan, Iran. The $x, y, z$ size of the obtained volumes was 512 × 650 × 128 voxels, 7 × 3.125 × 3.125 mm³, voxel size 13.67 × 4.81 × 24.41 $\mu m^3$(Fig.19.).

The OCT images were segmented to locate all of the 10 layers (11 surfaces) as given in [Quellec et al. 2010]. Furthermore, one extra surface was identified between the 6$^{th}$ and 7$^{th}$ surfaces (Fig.5). The important point in layer detection when employing this algorithm is that the number of clusters in the second diffusion map can be selected by the user and this makes the algorithm compatible with different types of OCT images which may not depict all expected surfaces. The validation was based on manual tracing by two observers. Our algorithms in their 2D and 3D versions were applied to this set and the mean signed and unsigned border positioning errors were computed for each surface and presented in TABLES VI and VII. Fig.28 shows examples of the achieved results.

TABLE VI
MEAN UNSIGNED BORDER POSITIONING ERRORS ON THE ISFAHAN DATASET (MEAN ± SD) IN MICROMETERS

| Border | Avg. Obs. Vs. Alg. in 2D approach | Avg. Obs. Vs. Alg. in 3D approach | Obs. 1 vs. Obs. 2 |
|---|---|---|---|
| 1 | 6.88±3.22 | 7.53±3.21 | 6.25±3.12 |
| 2 | 12.82±5.56 | 10.97±5.15 | 15.94±8.94 |
| 3 | 10.94±4.13 | 6.79±2.12 | 10.63±6.19 |
| 4 | 15.03±5.12 | 12.87±4.95 | 16.25±9.27 |
| 5 | 10.05±3.74 | 8.53±2.64 | 12.82±5.36 |
| 6 | 13.75±3.18 | 12.82±5.27 | 12.53±4.89 |
| 6a | 6.57±2.13 | 4.97±1.23 | 7.81±3.76 |
| 7 | 3.44±1.01 | 4.38±1.37 | 5.63±2.23 |
| 8 | 4.07±2.08 | 4.68±1.39 | 5.61±2.58 |
| 9 | 5.05±1.29 | 5.32±2.24 | 8.42±3.95 |
| 10 | 6.25±2.37 | 5.31±1.36 | 6.55±3.84 |
| 11 | 6.88±2.42 | 6.57±2.84 | 7.19±2.45 |
| overall | 8.52±3.13 | 7.56±2.95 | 9.65±4.83 |

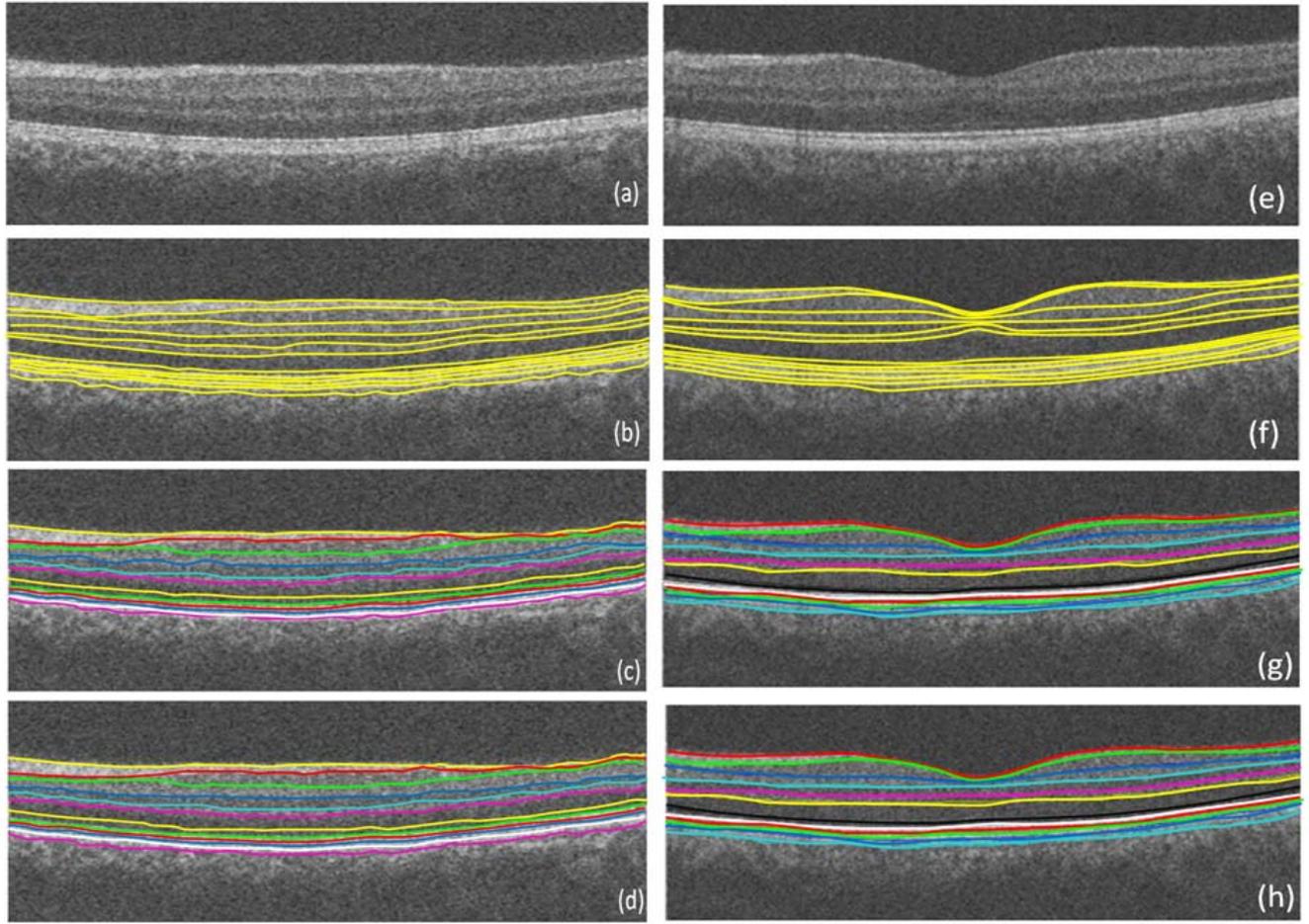

Fig.28. Two example results on the Isfahan dataset. (a,e) Composite image. (b,f) Composite image with independent standard (c,g). Composite image with segmented borders using the 2D approach. (d, h) Composite image with segmented borders using the 3D approach.

TABLE VII
MEAN SIGNED BORDER POSITIONING ERRORS (MEAN ± SD) IN MICROMETERS

| Border | Avg. Obs. Vs. Alg. in 2D approach | Avg. Obs. Vs. Alg. in 3D approach | Obs. 1 vs. Obs. 2 |
|---|---|---|---|
| 1 | 4.85±2.42 | 5.13±2.96 | -5.34±2.31 |
| 2 | -13.54±5.65 | -10.12±4.65 | 13.16±6.56 |
| 3 | -13.22±4.38 | -10.29±4.38 | 13.37±6.39 |
| 4 | -16.23±5.62 | -14.50±5.12 | 13.06±7.24 |
| 5 | -7.63±4.21 | -7.69±3.19 | 9.84±5.21 |
| 6 | -11.34±5.34 | -8.37±2.18 | 12.88±6.13 |
| 6a | 6.62±1.21 | -5.12±3.28 | -12.28±2.36 |
| 7 | 3.16±1.24 | 4.84±2.24 | -6.06±2.34 |
| 8 | 3.97±2.12 | 4.91±1.33 | 5.91±1.54 |
| 9 | -3.91±1.65 | -4.47±1.49 | 7.09±2.18 |
| 10 | -4.16±2.37 | -4.44±2.27 | 7.31±1.94 |
| 11 | -3.60±2.46 | -4.28±1.39 | 8.97±2.94 |
| overall | -4.61±3.35 | -4.53±2.89 | 5.71±3.98 |

As can be seen in TABLES VI and VII, the 1$^{st}$ and 7$^{th}$-to-11$^{th}$ boundaries, which are dependent on the gradient search, are better localized in the 2D approach. The reason was discussed in Section 4 (3D approach). To reduce the computational complexity, the gradient search was only applied to selected slices, which were located in borders of the determined cubes with a size of $10 \times 10 \times 10$. The boundaries of the remaining slices were then determined using simple interpolation (Fig.21, top). Therefore, it is expected that the border positioning error in the 3D approach will be greater than for the 2D method in the gradient-based boundaries. However, the localization error of other boundaries (labeled 2 – 6a) is lower for the 3D approach because of the global character of the 3D method.

### 5.3. A STEP BY STEP ALGORITHM FOR AUTOMATIC SEGMENTATION OF OCT DATASETS

As was presented in Section 3.3, the most influential outcome of this method is its ability to distinguish the proper number of layers in an OCT image. It should be mentioned that in some cases, very faint boundaries may not be distinguished by this method and two strategies may be developed: one is ignoring the missed layers, and the second method is setting a new diffusion map in the area of the missed layers. The second approach is applicable in missing boundaries, the anatomical existence of which are certain. Then, we may choose 2 boundaries positioned above and below the missed layer and remove the information out of this area (similar to operations performed in Figs.6 and 19). A new graph and diffusion map can then be employed to find the missed layers (as it is used in Section 5.2 to find the second layer (third boundary) and is demonstrated in Fig.29).

Following the "elbow" algorithm, k=6 was identified as the best cluster number (in general) for the second step of the diffusion map algorithm for our 2D OCT data. The cluster number was correct for the dataset used in Section 5.1 where the ELM/ISL interface wasn't clearly distinguishable and GCL/IPL interface was recognizable; however, using k=6 results in an incorrect segmentation in images, in which the ELM boundary exists explicitly while the GCL/IPL interface is very faded (like dataset used in section 5.2). In such cases, the existence of a clear boundary 6a, forces the algorithm to localize this boundary and miss a less visible boundary located between GCL and IPL. While we may apply k=7 (manually) and obtain correct localization of all layers (as demonstrated in Fig. 13), this is not an acceptable solution. Not only does it make our algorithm non-automated in detection of cluster numbers, but also this incorrect selection occurs in many cases. The solution is shown in Fig.29. As described above, two boundaries located above and below the known anatomical boundary may be selected in the next step and the diffusion map algorithm can be applied to the region located between them to find the missed edge.

To sum up, the steps for automatic computation of the number of clusters are as follows:

Step 1) Run the algorithm using the number of clusters obtained from the "elbow" algorithm.

Step 2) Calculate the mean vertical distances between the lowest localized boundary to two closest boundaries below and above.

Step 3)

    a. If the lowest boundary is nearer to the upper boundary than the lower one, we assume that the algorithm operates under normal conditions and that there is no need to take an extra action (similar to datasets used in section 5.1). Then go to Step 6.

    b. If, however, the lowest boundary is nearer to the lower boundary than the upper one, existence of a detectable ELM/ ISL boundary is assumed. Then go to Step 4.

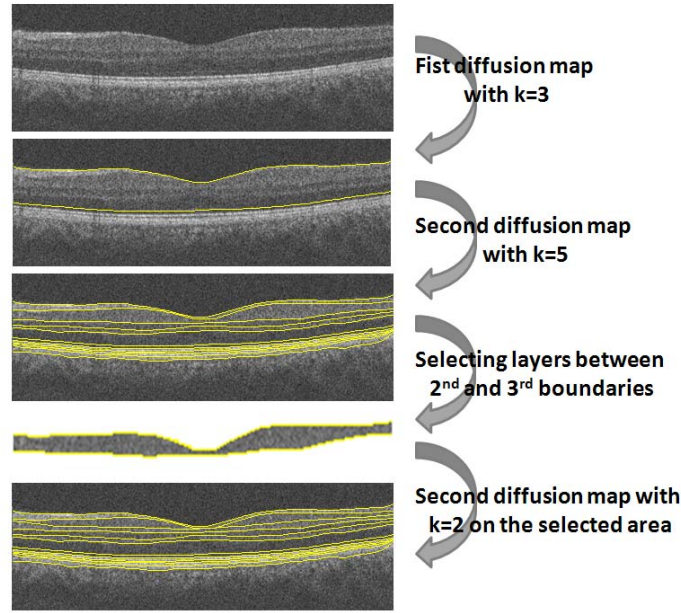

Fig. 29. Determining missed boundaries using successive diffusion maps.

Step 4)

    a. If the calculated number of clusters by the algorithm (using the eigen value method) is k=6, normal condition is assumed (with no merged layers due to pathology or imaging artifacts), but a faded GCL/IPL interface exists, then go to Step 5 (similar to datasets used in section 5.2).

    b. If the calculated number of clusters by the algorithm (using the eigen value method) is less than k=6, an abnormal condition is assumed with merged layers (examples of this condition can be found in Fig. 17). Then go to Step 6.

    c. If the calculated number of clusters by the algorithm (using eigen value method) is k=7, a normal condition is assumed (with no merged layers due to pathology or instrument), and the GCL/IPL interface is expected to be clear enough to be distinguished, then go to Step 6.

Step 5) Run another Diffusion map algorithm with k=2 in the area located between the second and third boundaries (as shown in Figure 29) to localize GCL/IPL interface. Then go to Step 6.

Step 6) Plot the boundaries and stop the algorithm.

The computation time of the proposed algorithms (implemented in MATLAB™ without using the mex files -Math Works, Inc., Natick, MA [MATLAB version 7.8]) were 12 seconds for each 2D slice and 230 seconds for each 3D volume using the 3D approach. The time complexity of the combined method (which used an interpolation on 2D results to estimate the 3D segmentation) was 1600 seconds. A PC with Microsoft Windows X x32 edition, Intel core 2 Duo CPU at 3.00GHz, 4 GB RAM was used for the 2D approach and Windows XP workstation with a 3.2GHz Intel Xeon CPU, 32 GB RAM was used for the 3D approach.

## 6. DISCUSSION AND CONCLUSION

A novel method for automatic segmentation of intraretinal layers in 2D and 3D SD-OCT scans from glaucoma patients and normal controls was presented. The method is based on application of two sequential diffusion maps, first of which segments the ILM-to-RPE complex and reduces the data to the region between these two layers. The second map then localizes the internal layers between the ILM and the RPE complexes.

The reported approach outperformed several previously published graph-based methods, which suggests an advantage of the employed texture analysis strategy.

Another important advantage of this method is its ability to make a correct decision regarding the number of clusters in k-means clustering, which, as a result, detects the proper number of layers to be segmented. This may be even more useful in cases for which one or more anatomical layers are not visible due to disease or low image quality.

The experiments suggested the robustness of this method across OCT scanners from different manufacturers, e.g. 3D spectral-domain OCT Zeiss Cirrus and Topcon 3D OCT-1000.

The computation time of this method is relatively low when compared with other methods especially considering its MATLAB implementation.

The robustness of the algorithm in presence of blood vessel artifacts is discussed in many papers [Yang et al. 2010], [Hood 2008], and some researchers tried to propose a preprocessing algorithm to compensate for the effect of these vessels [Shijian et al. 2010]]. However, the global approach behind the diffusion maps makes it intrinsically insensitive to small artifacts, as illustrated in Fig. 30.

Size of rectangles (or cubes) in sparse representation of the graphs is an important parameter of the approach. As described in Sections 3.2 and 4, in the 2D approach, the rectangle sizes of the first and second diffusion maps were set to $10 \times 10$ and $2 \times 20$, respectively. In the 3D approach, the size of cubes for the first diffusion map were selected as $10 \times 10 \times 10$ and the size of narrow prisms for the second diffusion map was set as $15 \times 1 \times 15$. As discussed above, the second diffusion map is applied on an aligned version of data, where we expect thin and smoothly spread surfaces for different layers. This can be a convincing reason for selecting thin rectangles (or prisms) in this step. In cases with retinal layer pathologies (which were not considered in this work), the surfaces may not be smooth and consequently, such thin rectangles (or cubes) may no longer be appropriate. Fig.31 shows an example of segmenting a pathologic dataset to demonstrate this problem. Fig.31(a,d) shows slices 125 and 135 (10 slice interval), located on the borders of the 3D cubes where the algorithm adjusts the $1^{st}$ and $7^{th}$ boundaries with a gradient search. The $130^{th}$ slice (located inside the cube) does not undergo the gradient search and the $1^{st}$ and $7^{th}$ boundaries are found using an interpolation on two sided slices numbered 125 and 135. Fig.31(b) shows a result of such interpolation which lacks accuracy. As a solution, one may perform the gradient search on every slice (which requires around 6 times longer processing time than the previous method) to obtain an acceptable result like that shown in Fig.31(c).

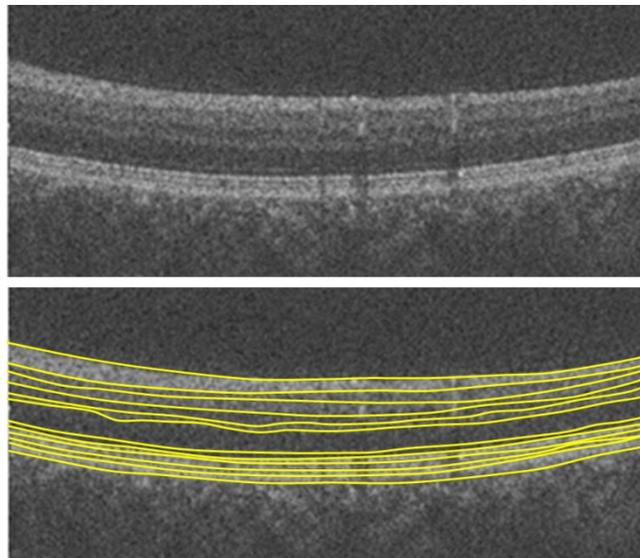

Fig. 30. Robustness of the proposed algorithm to blood vessel artifacts.

Therefore, it seems better to select a smaller number of pixels (or voxels) to construct the nodes. However, this assumption is not totally applicable since if (as an extreme) we select one pixel (or voxel) as a graph node, we lose the important advantage of noise cancellation obtained by the employed simple averaging of pixels. Therefore, the number of allocated pixels (or voxels) per node should be a tradeoff between the desired accuracy and needed noise management in the proposed method.

Another important point to be discussed is the advantage of employing diffusion maps in obtaining an initial rough approximation of the two initially detected boundaries (boundaries 1 and 7). We can compare our method to [Hee et al. 1995, Shahidi et al. 2005, Ishikawa et al. 2005, Koozekanani et al. 2001, Fabritius et al. 2000], which used a similar A-Scan search for finding increases of the noise level to identify ILM and PRE or looked for 1D edges and their maxima in each A-Scan. The most serious problem with these methods was identification of spurious responses. In contrast, the gradient search in our method is limited to an area around a rough approximation of these boundaries. Therefore, the speed of localization can be improved and the possibility of finding erroneous responses decreases. To have a numerical comparison, we developed the algorithm proposed by Fabritious [Fabritius et al. 2000] (to be one of the fastest methods heretofore) and the computation time of detecting boundaries 1 and 7 using the our algorithm and detection of ILM and PRE using [Fabritius et al. 2000] (on our 3D datasets used in section 5.2) were comparable for the two methods and both needed around 50 seconds for completion, demonstrating comparable speed of the two method.

Furthermore, to find the rough approximation, Ghorbel [Ghorbel 2011] proposed a k-means clustering on OCT image which didn't produce a discriminative segmentation (Figure 32) and the authors defined a great deal of extra steps to extract the boundaries. Our proposed method, however, performs the clustering in frequency domain and a meaningful segmentation is achieved simply requiring a subsequent gradient search to produce a final segmentation. The time complexity is also low since the image resolution is reduced by employing small regions in the sparse representation.

In our future work, we will focus on applications of diffusion wavelets [Coifman Maggioni 2006] on OCT images and will investigate their possible ability for noise removal, segmentation and detection of abnormalities of retinal OCT images.

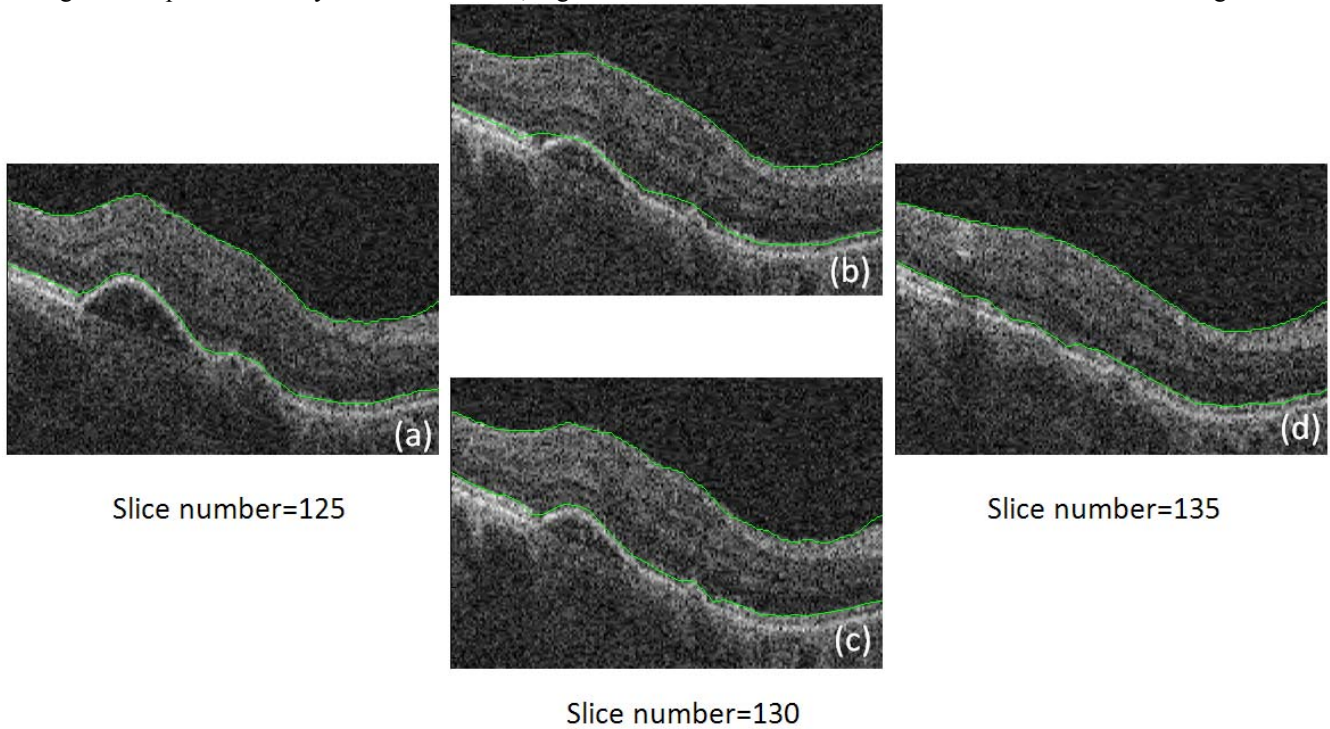

Fig. 31. Example of applying our approach to pathologic data. (a) Slice number 125, curves obtained by individual gradient search. (b) Slice number 130, interpolated curves. (c) Slice number 130, curves obtained by individual gradient search. (d) Slice number 135, curves obtained by individual gradient search.

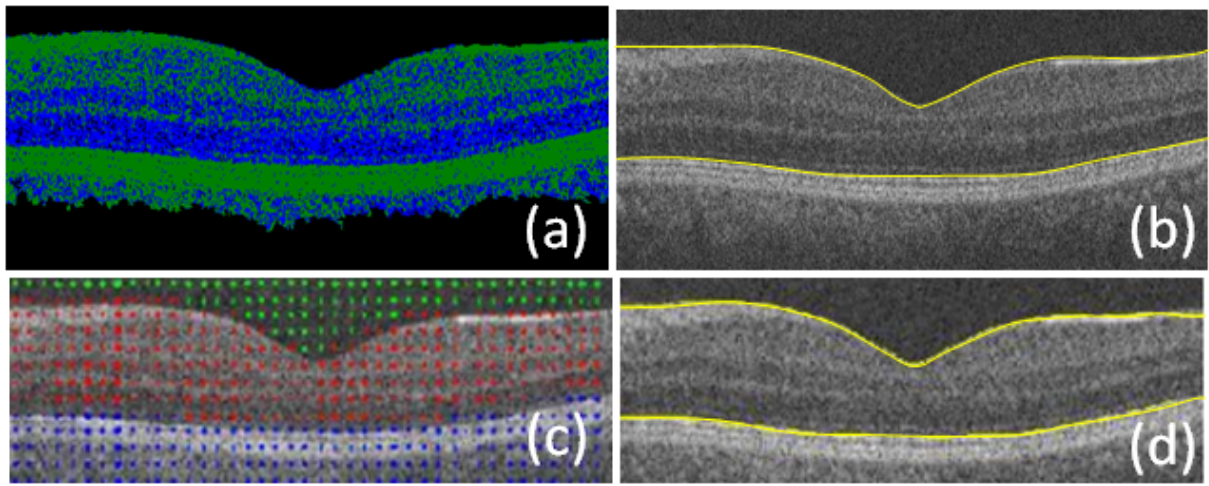

Fig. 32. A comparison of clustering in [Ghorbel 2011] and in the proposed method. (a) Clustering result obtained using Ghorbel's method [Ghorbel 2011]. (b) Segmentation resulting from the Ghorbel's method [Ghorbel 2011]. (c) Clustering obtained using the reported method. (d) Segmentation resulting from the proposed method.

## ACKNOWLEDGEMENTS

The authors would like to thank Dr. K. Lee and Mrs. B. Antony for preparation of data and/or results of their methods. This work was supported in part by the National Institutes of Health grants R01 EY018853, R01 EY019112, and R01 EB004640.